\documentclass[10pt,twocolumn,letterpaper]{article}

\usepackage{wacv}              

\usepackage[acronym]{glossaries}
\makeglossaries
\newacronym{CAD}{CAD}{Computer-Aided Design}
\newacronym{LLM}{LLM}{Large Language Models}
\newacronym{vlm}{VLM}{Vision Language Model}
\newacronym{vfm}{VFM}{Vision Foundation Model}
\newacronym{B-rep}{B-rep}{boundary representations}
\newacronym{MRR}{MRR}{mean reciprocal rank}

\usepackage{placeins}
\usepackage{graphicx}
\usepackage{subcaption}
\usepackage{float}
\usepackage{amsmath}
\usepackage{comment}
\usepackage{multirow}
\usepackage[table]{xcolor}
\usepackage{pdflscape}
\usepackage{makecell}
\usepackage[most]{tcolorbox}
\usepackage{gensymb}
\usepackage{siunitx}
\usepackage{cuted}   
\usepackage{caption}
\usepackage{svg}

\newacronym{VLMs}{VLMs}{Vision-Language Models}
\newacronym{LLMs}{LLMs}{Large Language Models}

\newtcolorbox{researchquestion}[1][VLM Prompt]{
  enhanced,
  breakable,
  colback=blue!3,
  colframe=blue!45!black,
  boxrule=0.5pt,
  arc=2pt,
  left=7pt,
  right=7pt,
  top=5pt,
  bottom=5pt,
  before skip=8pt,
  after skip=8pt,
  title=#1,
  fonttitle=\bfseries,
  coltitle=black,
  attach boxed title to top left={
    xshift=6pt,
    yshift=-2mm
  },
  boxed title style={
    colback=white,
    colframe=blue!45!black,
    boxrule=0.5pt,
    arc=1pt
  }
}

\definecolor{wacvblue}{rgb}{0.21,0.49,0.74}
\usepackage[pagebackref,breaklinks,colorlinks,allcolors=wacvblue]{hyperref}

\def\wacvPaperID{20} 
\def\confName{WACV}
\def\confYear{2027}

\title{Blueprint2CAD: Benchmarking Foundation Models on Engineering Drawing–to–CAD Render Correspondence}

\author{
Alexander Valverde$^{1,\dagger}$ \quad
Danny Xie-Li$^{2}$ \quad
Zhichao Wang$^{1}$ \quad
Marcos V. Conde$^{3}$ \\[0.5em]
$^{1}$ Novineer \quad
$^{2}$ Instituto Tecnol\'ogico de Costa Rica \quad
$^{3}$ University of W\"urzburg \\[0.3em]
$^{\dagger}$ Corresponding author: {\tt\small alexander.valverde@novineer.com}
}

\begin{document}

\twocolumn[{%
\renewcommand\twocolumn[1][]{#1}%
\maketitle

\vspace{-7.5mm}
\begin{center}
    \includegraphics[width=0.95\linewidth]{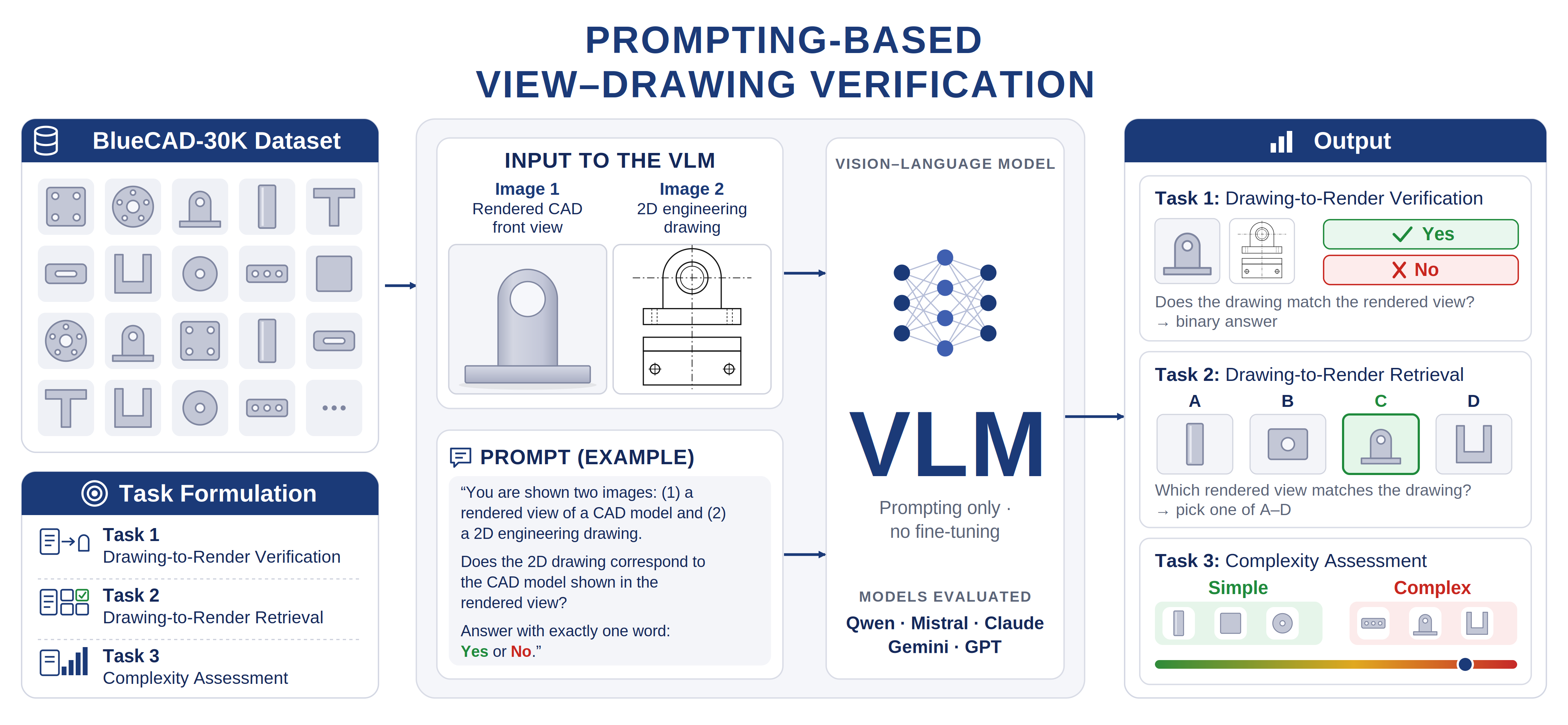}
    \vspace{-2.3mm}
    \captionof{figure}{Overview of Blueprint2CAD. All tasks pair two 2D images, an engineering line drawing and a render of a part.}
    \label{fig_teaser_image}
    \vspace*{2.7mm}
\end{center}
}]

\begin{abstract}
Engineering drawings specify mechanical parts through orthographic projections. Recognizing which part a drawing depicts underlies part search, design reuse, and digitization, yet no public benchmark measures whether vision models can do it. We introduce \textbf{Blueprint2CAD}, a benchmark for drawing-to-render correspondence: given an orthographic line drawing and a shaded render, usually from another viewpoint, decide whether they show the same part. Its data layer, \textbf{BlueCAD-30K}, applies an open drafting pipeline to 29,925 Zero-to-CAD models, adding view-aligned renders, clean and dimensioned drawings from six viewpoints, and surface-type metadata. Frozen encoders retrieve the matching part from a 10,000-model gallery with up to 65\% Recall@1 when viewpoints are known but 43\% when they are not, and the four encoders we test lie within six points of each other: viewpoint, not backbone, is the bottleneck. Six zero-shot vision-language models reach only 0.52--0.67 verification accuracy, with model-specific \emph{Yes}/\emph{No} biases that chain-of-thought shifts but does not remove; a quick LoRA fine-tuning on in-domain pairs improves verification accuracy by a few points but sharpens these biases and leaves the viewpoint failure intact. \footnote{Code and benchmark are available at:  \url{https://github.com/alexvalgui/blueprint2cad}.}
\end{abstract}    
\section{Introduction}
Engineering drawings are how mechanical parts are specified, archived, and exchanged. A drawing encodes shape implicitly, through orthographic projections and drafting conventions~\cite{iso128,giesecke2016technical}, while the part itself exists as an exact CAD boundary representation (B-rep)~\cite{requicha1980representations,mantyla1988solid}. Recognizing which part a drawing depicts is the first step in part-library search, legacy-design reuse, drawing digitization, and model verification, and any automated pipeline from drawings to solids depends on a visual front-end that does this reliably.

The problem is harder than it looks. Mechanical components are textureless, symmetric, and repetitive, and their identity rests on small features such as holes, pockets, fillets, and chamfers. A drawing presents these as sparse line art with hidden lines and annotations layered on top, whereas a render shows a shaded surface, usually from another viewpoint. Deciding whether the two show the same part therefore means holding one projection and relating it to another across a change of view, a correspondence that is natural to carry as visual structure and awkward to spell out in words. Existing CAD datasets~\cite{koch2019abc,kim2020mcb,willis2021fusion360,manda2021cadsketchnet} support shape retrieval and generative modeling, and SPARE3D~\cite{han2020spare3d} poses spatial-reasoning questions over three-view drawings, but none pairs engineering drawings with shaded renders of the same part under controlled cameras, so the encoders behind drawing-to-CAD reconstruction~\cite{wang2024cad2program,dimitrov2023engineering}, drawing digitization~\cite{moreno2021computer,de2022symbol}, and CAD inspection~\cite{alrashedy2025cadcodeverify} have never been measured on orthographic line art.

We introduce \textbf{Blueprint2CAD}, a benchmark that makes drawing-to-render correspondence a first-class task. Its scope is deliberately 2D: every task relates a line drawing to an image of the part it describes, and nothing is claimed about recovering 3D geometry. The data layer, \textbf{BlueCAD-30K}, adds no new geometry either. It is the output of an open drafting pipeline applied to 29,925 models of the Zero-to-CAD collection~\cite{ataei2026zerotocadagenticsynthesisinterpretable}: shaded renders together with clean and dimensioned projections from six canonical viewpoints under a shared camera rig, so that view correspondence holds by construction, plus B-rep-derived surface-type metadata and model-level training, validation, and test splits. The pipeline runs on any STEP collection, which is how we expect the benchmark to grow.

We evaluate the two paradigms a practitioner would reach for first. Four frozen vision foundation models (DINOv3~\cite{simeoni2025dinov3}, SigLIP2~\cite{tschannen2025siglip2multilingualvisionlanguage}, RADIO~\cite{Ranzinger_2024_CVPR}, TIPSv2~\cite{cao2026tipsv2}) serve as embeddings for search over a 10,000-model gallery: when the camera direction of each projection is known, the four strongest reach 59--65\% Recall@1 within six points of each other, while a hand-crafted silhouette descriptor already reaches 49\%; without that knowledge the best encoder falls to 43\%, still about twice the silhouette's rate, two dimension lines per view cost at most seven points, and I-JEPA stays near chance under our protocol. Six vision-language models (Claude Sonnet~5~\cite{anthropic2026sonnet5}, Gemini~3.6~Flash~\cite{gemini36flash}, GPT-5.6~Luna~\cite{openai2026gpt56}, Mistral Small~3.2~24B~\cite{mistralsmall2025}, InternVL3.5~14B~\cite{wang2025internvl35advancingopensourcemultimodal}, Qwen3-VL~8B~\cite{bai2025qwen3vltechnicalreport}) are queried zero-shot on binary verification and 4-way retrieval: none reaches 0.70 verification accuracy, even with chain-of-thought, each carries its own \emph{Yes}/\emph{No} prior that reasoning prompts redirect rather than remove, and correct matches become rare once the render is a top or bottom view. A tiny LoRA adapter trained on in-domain pairs lifts Qwen3-VL~8B by a few points without touching either failure mode. Read together, the two studies place drawing-to-render correspondence among the problems where an intermediate representation in language does not stand in for one that keeps the geometry: supplying the view relation directly, as the oracle alignment does for frozen encoders, is worth more than any change of backbone, while asking a \gls{vlm} to narrate that relation leaves its viewpoint failure where it was. Our contributions are:
\begin{itemize}[leftmargin=*,nosep]
    \item \textbf{Blueprint2CAD}, a 2D benchmark for matching engineering drawings to part renders, covering gallery-scale retrieval, two direct \gls{vlm} tasks, and an unguided complexity probe.
    \item An open drafting pipeline for any STEP collection, and \textbf{BlueCAD-30K}, the line art, shaded images, metadata, and model-level splits it adds to 29,925 Zero-to-CAD models.
    \item A \textbf{diagnostic study} of the five encoders against a non-learned silhouette baseline, under view alignment, multi-view fusion, and annotation clutter, and a failure-mode analysis of the six \gls{vlm}s and one fine-tuned variant that isolates model-specific response biases and tests whether reasoning in language recovers the cross-view correspondence the task requires.
\end{itemize}
\section{Related Work}

\paragraph{Reasoning with visual intermediates.}
A model can externalize an intermediate as pixels or tokens~\cite{su2025thinkingwithimages,hu2024visualsketchpad,li2025mvot,bigverdi2025perceptiontokens}, keep it latent in the encoder's embedding space~\cite{yang2025mirage,li2025lvr,xiao2026realvr}, or imagine the unseen view with a world model or spatial map~\cite{yang2025mindjourney,wang2025mindcube}, because textual chain-of-thought does not improve and can degrade spatial judgments~\cite{yang2024thinkinginspace,kancheti2026cotdegrades}, binary verification follows a model-specific \emph{Yes} prior~\cite{chan2026yesbias}, and \glspl{vlm} default to the camera's frame under viewpoint change~\cite{zhang2025spinbench,zhan20263viewsensespatialmentalperspective,michalkiewicz2026giqbenchmarking3dgeometric}. None tests whether an intermediate carries an orthographic drawing to a render of the same part; Blueprint2CAD measures that, and its view labels make these methods testable on it.

\paragraph{Engineering drawings}
Drawings have long been treated as a document-analysis problem: vectorization of scans, symbol and dimension parsing~\cite{moreno2021computer,de2022symbol}, and recently drawing-centric VQA benchmarks for multimodal models~\cite{kou2026mechvqabenchmarkingenhancingmultimodal,niu2025creftcadboostingorthographicprojection}, with reconstruction from orthographic views~\cite{idesawa1973system,hu2023plankassembly,wang2024cad2program,qin2025drawing2cad} as the downstream goal. SPARE3D~\cite{han2020spare3d} comes closest, asking view-consistency and camera-pose questions over three-view line drawings, but both sides of each question are line art. Missing is a measurement of the step in between: whether a general-purpose encoder or \gls{vlm} can read a drawing well enough to recognize the part it depicts.

\paragraph{2D--3D CAD retrieval and foundation models.}
Image-to-CAD retrieval learns joint embeddings of images and 3D geometry, from RGB-D scans~\cite{Avetisyan_2019_CVPR} to monocular RGB~\cite{kuo2020mask2cad,kuo2021patch2cad,gumeli2022roca,gao2024diffcad,maninis2023cadestate}, and sketch-based retrieval addresses the line-drawing-to-shape gap on SHREC~\cite{dai2017correlated,chen2018sketch3d,lei2019p2s,li2013shrec,qin2022shrec}; both target household objects with dedicated 3D encoders. Frozen encoders such as CLIP~\cite{radford2021clip}, DINO~\cite{oquab2023dinov2,simeoni2025dinov3}, SAM~\cite{kirillov2023sam}, and RADIO~\cite{Ranzinger_2024_CVPR} are distilled into 3D encoders by ULIP~\cite{xue2023ulip,xue2024ulip2}, OpenShape~\cite{liu2023openshape}, and Uni3D~\cite{zhou2024uni3d}, but their natural-image pretraining is not tuned to untextured, symmetric mechanical parts; CADCrafter~\cite{chen2025cadcrafter} conditions on depth and normals instead, and a growing line fine-tunes multimodal \gls{LLMs} to emit parametric CAD directly~\cite{yuan2024openecad,xu2024cadmllm,wang2025cadgpt,wu2024cadvlm,rukhovich2025cadrecode,chen2024img2cad,you2024img2cad}. We instead measure purely 2D encoders with a line drawing as the query and a render of an engineering part as the target.

\paragraph{VLMs and native CAD learning.}
\gls{vlm} work on CAD covers generation from visual input~\cite{kolodiazhnyi2025cadrille,doris2025cadcoderopensourcevisionlanguagemodel,joglekar2026ortho2cad3dcadgeneration}, drawing VQA where correspondence is a small subtask~\cite{kou2026mechvqabenchmarkingenhancingmultimodal,niu2025creftcadboostingorthographicprojection}, and embedding-space sketch matching~\cite{hou2024sketch}; none poses identity verification between an engineering drawing and a shaded render as a primary task on a public dataset. Separately, generative models emit sketch-and-extrude sequences~\cite{wu2021deepcad,xu2022skexgen,xu2023hnccad}, B-rep encoders classify and retrieve parts~\cite{jayaraman2021uvnet,lambourne2021brepnet}, and reverse engineering recovers models from point clouds~\cite{uy2022point2cyl,liu2024point2cad,khan2024cadsignet}, sketches~\cite{li2022free2cad,seff2022vitruvion}, or drawings~\cite{hu2023plankassembly,wang2024cad2program,qin2025drawing2cad}, supported by datasets with design histories~\cite{willis2021fusion360,seff2020sketchgraphs,khan2024text2cad}. We use B-rep machinery only to describe and sample our data; B-rep-native encoders are left to future work.

\section{Background}

CAD systems~\cite{sutherland1963sketchpad,requicha1980representations} describe a solid as a B-rep: faces, edges, and vertices carried by parametric surfaces and curves~\cite{braid1975synthesis,mantyla1988solid}, with freeform geometry expressed as NURBS~\cite{piegl1997nurbs} and built as sequences of operations over constrained 2D sketches~\cite{shah1995parametric}. The CAD model is the primary definition from which engineering drawings are derived. A drawing follows Monge's descriptive geometry~\cite{monge1799geometrie}: a 3D object is represented exactly by its orthographic projections onto perpendicular planes, stored digitally as vector primitives (segments, arcs, circles, splines) interleaved with dimensions, text, and hatching in formats such as DXF~\cite{autodesk_dxf} or SVG~\cite{w3c_svg}. In practice drawings range from bare orthographic outlines to section and auxiliary views, GD\&T, surface-finish callouts, and title blocks~\cite{iso128,giesecke2016technical}; this work addresses the orthographic core shared by all of them (Sec.~\ref{sec:limitations}). Because the 3D geometry is encoded implicitly across complementary views, reconstructing a model from its drawing has been an open problem since the earliest work on orthographic reconstruction~\cite{idesawa1973system}.
\section{BlueCAD-30K}

\begin{figure*}[t]
    \centering
    \begin{tabular}{c c}
         \includegraphics[width=0.43\linewidth]{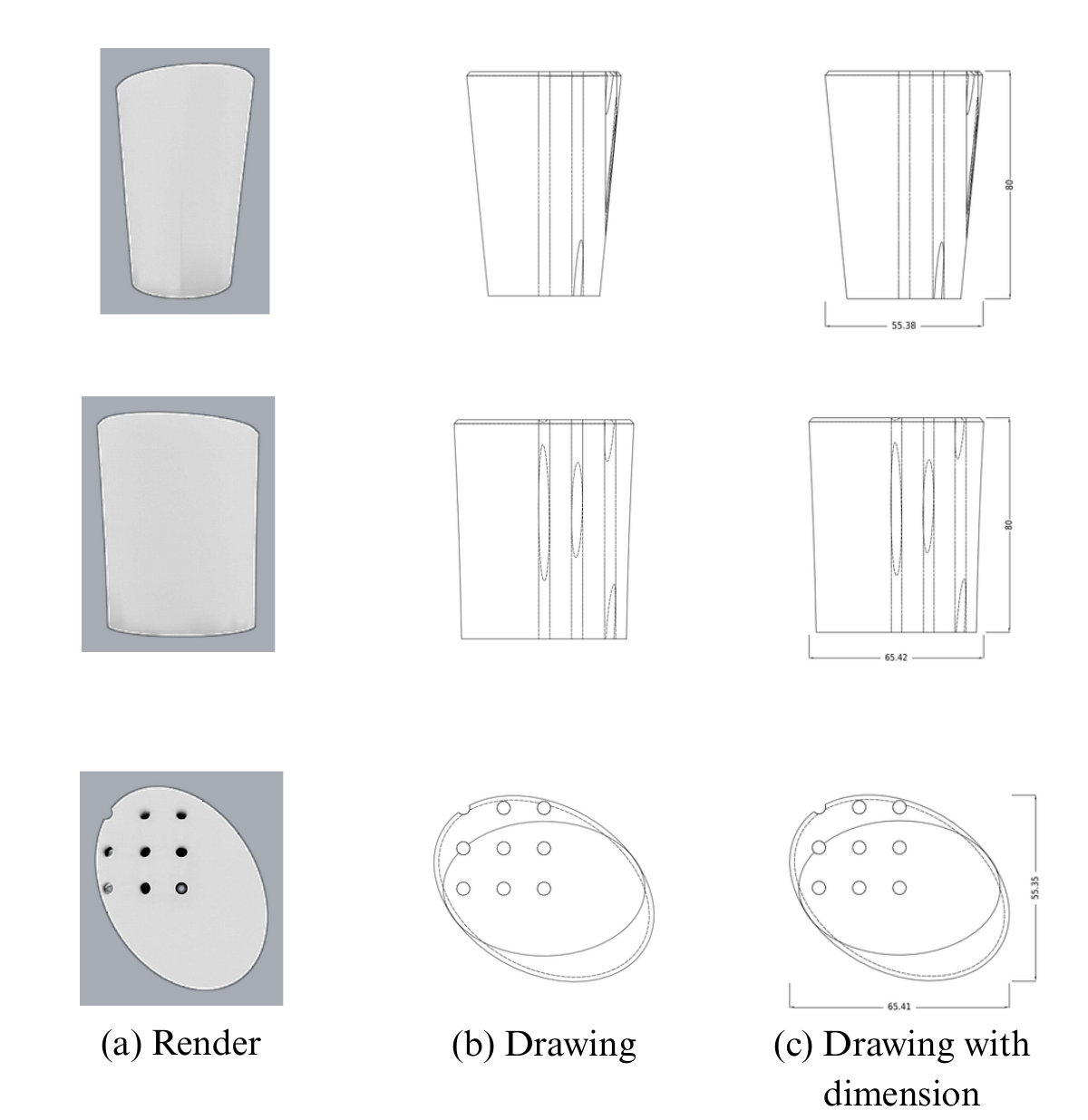} &
         \includegraphics[width=0.43\linewidth]{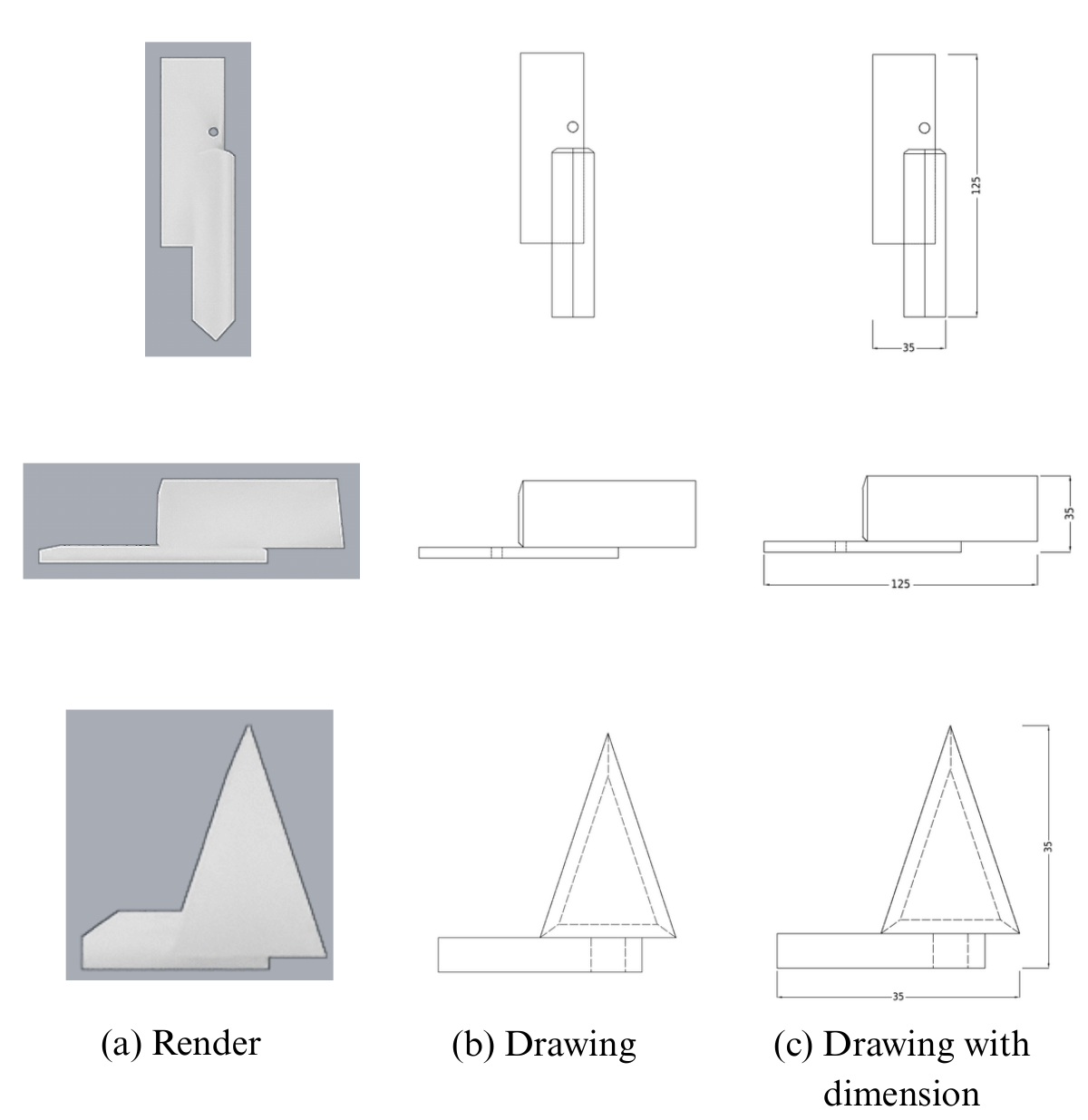} \\
         Freeform subset (\emph{complex}) & Prismatic subset (\emph{simple}) \\
    \end{tabular}
    \caption{\textbf{Examples of CAD models} from the two \gls{vlm} evaluation subsets: shaded render, clean 2D drawing, and dimensioned 2D drawing.}
    \label{fig:sample_grid_cad}
\end{figure*}

\subsection{Dataset}
\label{sec:dataset}

\paragraph{Origin.}
BlueCAD-30K is built on Zero-to-CAD~\cite{ataei2026zerotocadagenticsynthesisinterpretable}, a 2026 collection of STEP models generated from procedural CAD scripts. We add no geometry: the dataset is the drawings, renders, and metadata our pipeline produces for a filtered subset, with source models referenced by identifier. We chose Zero-to-CAD over scraped repositories~\cite{koch2019abc,willis2021fusion360,kim2020mcb} because paired drawing generation needs clean, watertight, single-solid geometry; the pipeline runs on any STEP collection and can be re-instantiated on real part libraries (Sec.~\ref{sec:limitations}).

\paragraph{Filtering.}
A B-rep parser removes models with a low geometry score or curved-surface ratio, degenerate aspect ratios, non-watertight geometry, or fewer than $k{=}5$ faces, reducing the 1M pool to roughly 300K. A random 10\% sample gives 30,237 models; 29,925 pass every stage, and 312 (1.0\%) fail in drafting or rendering because of invalid topology the parser misses. In the sample, 65\% of faces are planar and 22\% cylindrical, 24\% of models contain a B-spline face, and 0.02\% are exact topological duplicates; geometric diversity is bounded by the single generator (Sec.~\ref{sec:limitations}).

\begin{figure}
    \centering
    \includegraphics[width=1\linewidth]{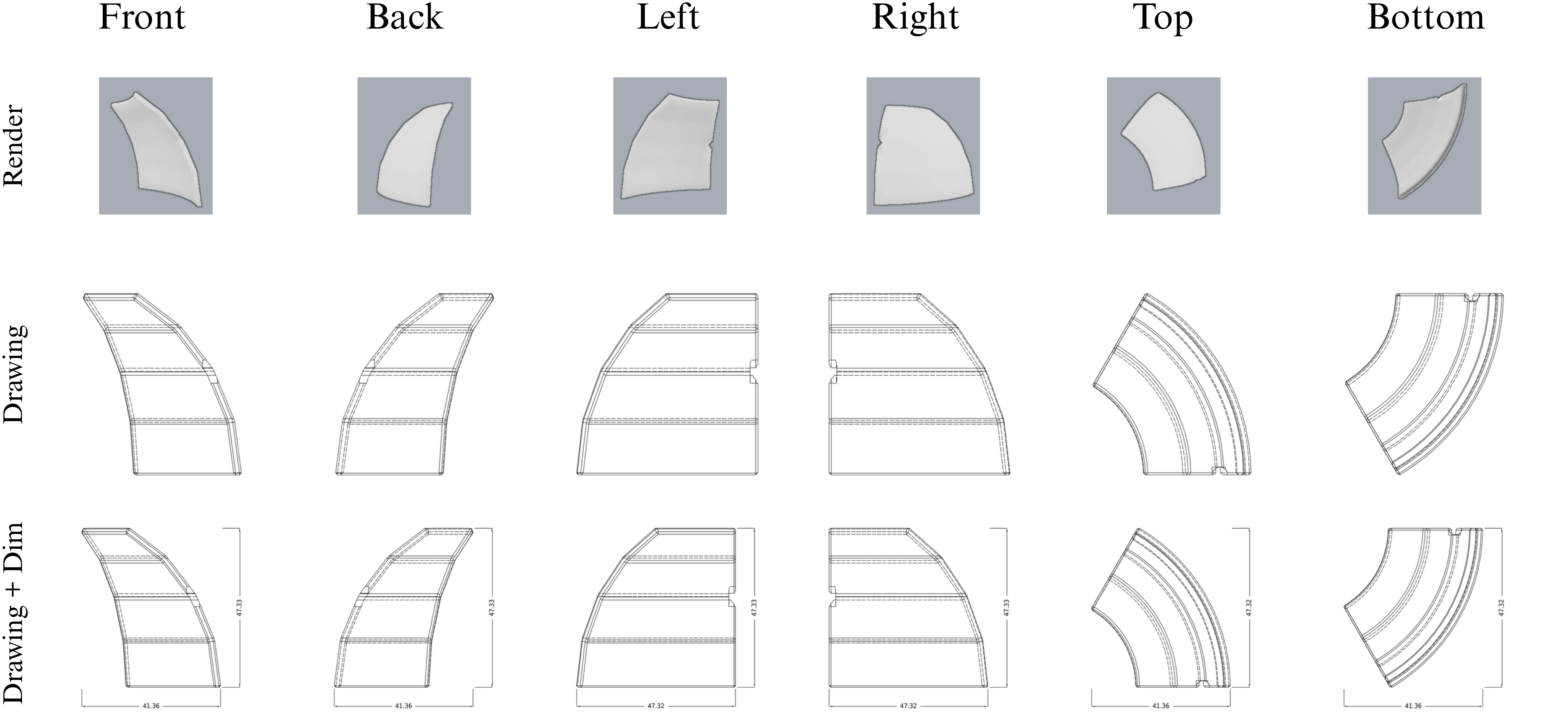}
    \caption{\textbf{A CAD example} with its six camera directions and paired render, drawing, and dimensioned drawing views.}
    \label{fig:camera_views}
\end{figure}

\paragraph{Drafting pipeline.}
Each model is projected in native STEP coordinates from six canonical orthographic viewpoints shared by drawing and render, so view correspondence holds by construction (Fig.~\ref{fig:camera_views}; axes and edge treatment in the supplementary). Drawings use OpenCascade hidden-line removal, are stored as layered DXF, and are rasterized to $1024\times1024$\,px on white; renders have the same resolution and a transparent background. A dimensioned variant adds width and height dimensions to each view. Data (CC~BY~4.0) and code (Apache~2.0) are released: per model, six renders, six clean and six dimensioned drawings, and B-rep metadata, plus the \gls{vlm} pairs, prompts, checkpoints, and raw responses.

\paragraph{Splits.}
Splits are made by model identifier, so no model appears in two of them: 27,479 models for training, 1,446 for validation (5\% of the non-test models, fixed seed), and 1,000 for testing, which form the \gls{vlm} evaluation set (Sec.~\ref{sec:visual_language_dataset}). Training pairs draw negatives only from their own split (supplementary Sec.~\ref{app:finetune}). The frozen-encoder study uses a 10,000-model gallery sampled from the dataset. All splits, including the gallery, are released.

\subsection{VLM Evaluation Set}
\label{sec:visual_language_dataset}
The \gls{vlm} evaluation set holds the 500 lowest- and 500 highest-scoring models under a selection score $s\in[0,100]$, a weighted sum of seven normalized B-rep signals (freeform-face ratio, face count, non-planar ratio, surface-type entropy, inner loops, curved-edge ratio, and sphericity; weights in supplementary Table~\ref{tab:complexity_signals}, following~\cite{arslan2021shrec,jayaraman2021uvnet,lambourne2021brepnet,wadell1935volume}). The low end consists of \emph{prismatic} parts with planar and cylindrical faces, the high end of \emph{freeform} parts dominated by B-spline surfaces. The score only selects the two sets, which do not overlap in score ($12.2$--$18.2$ vs.\ $59.1$--$87.3$): 99\% of prismatic models have no freeform face and 95\% of freeform models have a non-planar ratio above $0.6$, while face counts overlap (interquartile ranges 19--32 and 26--63), so the classes differ in surface type, not size (supplementary Sec.~\ref{app:stats}). Tables label them \emph{simple} and \emph{complex} to match the Task~3 prompt.
\providecommand{\tbd}[1]{\textcolor{red}{[TBD: #1]}}
\section{Blueprint2CAD Benchmark}
We examine drawing-to-render correspondence from two angles: frozen \gls{vfm}s as cross-modal embeddings for retrieval (Sec.~\ref{sec:visual_foundational_models}), and \gls{vlm}s queried directly on matching tasks (Sec.~\ref{sec:vision_language_models}). Retrieval runs over a 10{,}000-model sample of the full dataset, and the \gls{vlm} tasks over the 1{,}000-model evaluation set. Every input is a pair of 2D images, a line drawing on one side and a shaded render on the other. Models run frozen or zero-shot, with one exception, a LoRA-adapted Qwen3-VL~8B (Sec.~\ref{sec:vlm_findings}), a small-scale reference that exercises the training split (Sec.~\ref{sec:limitations}).

\subsection{Vision Foundation Models}
\label{sec:visual_foundational_models}
We ask whether frozen 2D \gls{vfm}s can link a drawing to the render of the same part, probing four backbones with distinct pretraining objectives: DINOv3~\cite{simeoni2025dinov3}, SigLIP2~\cite{tschannen2025siglip2multilingualvisionlanguage}, RADIO~\cite{Ranzinger_2024_CVPR}, and TIPSv2~\cite{cao2026tipsv2}. Freezing is deliberate: digitization and reconstruction pipelines use these encoders off the shelf, so we measure what pretraining already captures. Since nothing is learned, no training split is needed, and the gallery is a 10{,}000-model sample of the dataset. Each backbone yields one global descriptor per drawing and render view of every model, twelve in total (Sec.~\ref{sec:dataset}), mean-centered per modality over the gallery and $\ell_2$-normalized; checkpoints, resolutions, and descriptor choices are in the supplementary.

\paragraph{Cross-modal retrieval protocol.}
Each model thus has $V{=}6$ unit-normalized drawing descriptors $\hat d_i$ and render descriptors $\hat r_i$, indexed by viewpoint. A query in one modality must retrieve its counterpart among all 10{,}000 gallery models; we report bidirectional Recall@$k$ (R@$k$), \gls{MRR}, and median rank (chance: R@1 $= 0.01\%$, median rank $\approx 5{,}000$). With unknown viewpoints (\emph{view-agnostic}), \emph{Max} takes the highest similarity over all $V{\times}V$ pairs and \emph{Mean} compares the re-normalized per-model averages. With known viewpoints (\emph{view-aligned}), only matching projections are compared, $s_{\text{aligned}} = \operatorname{agg}_{i}\, \hat d_i^{\top}\hat r_i$, with $\operatorname{agg}$ the mean (\emph{Al-mean}) or the max (\emph{Al-max}). Since alignment uses oracle labels, it is a diagnostic of what a representation can do once viewpoint is solved; the view-agnostic setting is the deployable one.

\paragraph{Experiment 1: Frozen Backbones.}
Tab.~\ref{tab:exp1} separates two factors. The backbone matters little: DINOv3, TIPSv2, SigLIP2, and RADIO, pretrained by self-distillation, image--text contrast, and multi-teacher distillation, finish within six R@1 points of each other under Al-mean. Viewpoint correspondence matters far more, adding 22--30 points over each encoder's best view-agnostic score, so the deployable setting is limited by viewpoint rather than by representation quality. Outline alone explains much of this: the silhouette baseline reaches three quarters of the encoders' aligned R@1 and gains as much from alignment. The encoders lead it by 11--17 points when viewpoints are known but retrieve 1.7--2.1 times as often when they are not, so pretraining mainly adds a view-consistent description of the part rather than a reading of its interior. Averaging over views beats the best single view in both settings, since several weak projections outweigh one that different parts may share. All four encoders retrieve better from drawings to renders than the reverse while the silhouette baseline is symmetric, placing the asymmetry in how the encoders embed line art and shaded renders. DINOv3 has the best aligned R@1 and MRR and is the default below.

\begin{table}[t]
\centering
\caption{\textbf{Experiment 1: frozen-backbone cross-modal retrieval.} R@1 (\%, D$\rightarrow$R) for the four scoring strategies defined in the protocol: \emph{Max}/\emph{Mean} are view-agnostic, \emph{Al-max}/\emph{Al-mean} assume known viewpoints. MRR, median rank, and reverse direction for Al-mean. \emph{Silhouette}: non-learned outline baseline, six filled $16{\times}16$ view masks plus aspect ratio (the hard-negative descriptor of Sec.~\ref{app:finetune}). Bold: best per column. Full metrics in the supplementary.}
\label{tab:exp1}
\resizebox{\linewidth}{!}{%
\begin{tabular}{lcc|cc|ccc}
\toprule
\rowcolor{gray!20}
& \multicolumn{2}{c|}{\textbf{R@1}, view-agnostic $\uparrow$} & \multicolumn{2}{c|}{\textbf{R@1}, view-aligned $\uparrow$} & \multicolumn{3}{c}{\textbf{Al-mean}} \\
\rowcolor{gray!20}
\textbf{Backbone} & \textbf{Max} & \textbf{Mean} & \textbf{Al-max} & \textbf{Al-mean} & \textbf{R@1 (R$\rightarrow$D)} $\uparrow$ & \textbf{MRR} $\uparrow$ & \textbf{MedR} $\downarrow$ \\
\midrule
Silhouette & 7.85 & 20.16 & 10.96 & 48.54 & 48.08 & 0.555 & 2 \\
\midrule
DINOv3  & 29.10 & \textbf{43.11} & 34.67 & \textbf{65.14} & 57.57 & \textbf{0.728} & \textbf{1} \\
TIPSv2  & \textbf{34.39} & 34.65 & \textbf{38.58} & 64.65 & \textbf{58.67} & 0.724 & \textbf{1} \\
SigLIP2 & 29.52 & 38.07 & 33.54 & 64.15 & 57.15 & 0.714 & \textbf{1} \\
RADIO   & 27.46 & 36.54 & 30.98 & 59.51 & 49.58 & 0.675 & \textbf{1} \\
\bottomrule
\end{tabular}}
\end{table}

\paragraph{Experiment 2: Per-View Analysis.}
Tab.~\ref{tab:exp2} adds views to DINOv3 one at a time. Gains come from new viewing axes, not from more pictures: \emph{back} adds under a point to \emph{front}, and for all four encoders an opposing pair improves on its stronger member by at most 2.5 points (supplementary), since the two projections mirror each other. Plan views carry the most information: \emph{top} and \emph{bottom} are the strongest single views for every encoder, roughly twice any side view, and top\,+\,left is every encoder's best pair, plausibly because extruded mechanical parts expose holes and pockets in plan that side views reduce to lines. The late jump at \emph{top} follows from this ranking, as it outweighs the weaker \emph{bottom} view it mirrors; the sixth view then changes R@1 by under half a point for every encoder.

\begin{table}[t]
\centering
\caption{\textbf{Experiment 2: cumulative view fusion} (frozen DINOv3, Al-mean, D$\rightarrow$R). Gains concentrate where a new viewing axis is introduced and on the top view; the opposing back and right views add almost nothing. Single- and two-view results for all encoders in the supplementary.}
\label{tab:exp2}
\resizebox{\linewidth}{!}{%
\begin{tabular}{lccccc}
\toprule
\rowcolor{gray!20}
\textbf{Views} & \textbf{R@1}$\uparrow$ & \textbf{R@5}$\uparrow$ & \textbf{R@10}$\uparrow$ & \textbf{MRR}$\uparrow$ & \textbf{MedR}$\downarrow$ \\
\midrule
front             & 14.88 & 25.98 & 31.38 & 0.206 & 62 \\
\ + back          & 15.76 & 27.08 & 33.11 & 0.217 & 48 \\
\ + bottom        & 42.19 & 61.12 & 68.20 & 0.512 & 2 \\
\ + left          & 52.15 & 70.74 & 76.93 & 0.607 & \textbf{1} \\
\ + top           & \textbf{65.59} & \textbf{81.90} & \textbf{87.15} & \textbf{0.731} & \textbf{1} \\
\ + right (all 6) & 65.14 & 81.79 & 86.87 & 0.728 & \textbf{1} \\
\bottomrule
\end{tabular}}
\end{table}

\begin{table}[t]
\centering
\caption{\textbf{Experiment 3: annotation ablation} (normal\,$\rightarrow$\,dim, Al-mean, D$\rightarrow$R). Worse / Better: share of queries whose rank degrades / improves with dimension annotations; the rest keep their rank. Bold: best value before and after annotation. Full metrics in the supplementary.}
\label{tab:exp3}
\resizebox{\linewidth}{!}{%
\begin{tabular}{lccccc}
\toprule
\rowcolor{gray!20}
\textbf{Backbone} & \textbf{R@1} (\%) $\uparrow$ & \textbf{MRR} $\uparrow$ & \textbf{MedR} & \textbf{Worse} (\%) & \textbf{Better} (\%) \\
\midrule
DINOv3  & \textbf{65.14} $\rightarrow$ 58.10 & \textbf{0.728} $\rightarrow$ 0.663 & 1 $\rightarrow$ 1 & 29.1 & 13.4 \\
TIPSv2  & 64.65 $\rightarrow$ \textbf{61.61} & 0.724 $\rightarrow$ \textbf{0.697} & 1 $\rightarrow$ 1 & 23.2 & 16.0 \\
SigLIP2 & 64.15 $\rightarrow$ 61.51 & 0.714 $\rightarrow$ 0.692 & 1 $\rightarrow$ 1 & 23.7 & 20.3 \\
RADIO   & 59.51 $\rightarrow$ 58.84 & 0.675 $\rightarrow$ 0.673 & 1 $\rightarrow$ 1 & 22.5 & 22.9 \\
\bottomrule
\end{tabular}}
\end{table}

\paragraph{Experiment 3: Drawing Annotations.}
Real engineering drawings carry dimension annotations (extension lines, arrows, measurement text) with no counterpart in a render. Our dimensioned variant (Tab.~\ref{tab:exp3}) adds two callouts per view, far fewer than a production drawing carries, and its cost is small next to the viewpoint problem: at most 7 R@1 points against 22--30 for unknown viewpoints. Annotations reorder close candidates rather than lose the match: more than half of the queries keep their exact rank, the rest move by a median of 4--5 positions, and every gap narrows from R@1 to R@10 (supplementary). Sensitivity differs by backbone. DINOv3, the only one of the four trained without language supervision, loses most and drops from first to last, while RADIO's change is not significant (sign test $|z| = 0.5$). With four encoders this link to language supervision is a correlation, and whether the measurement text or the extension lines drive it remains untested.

\subsection{Vision-Language Models Benchmark}
\label{sec:vision_language_models}
We evaluate six \gls{vlm}s, three closed-source (Gemini 3.6 Flash \cite{gemini36flash}, GPT-5.6 Luna \cite{openai2026gpt56}, Claude Sonnet 5 \cite{anthropic2026sonnet5}) and three open-source (Mistral Small 3.2 24B \cite{mistralsmall2025}, Qwen3-VL 8B \cite{bai2025qwen3vltechnicalreport}, InternVL3.5 14B \cite{wang2025internvl35advancingopensourcemultimodal}), zero-shot on the $1{,}000$-model evaluation set of Sec.~\ref{sec:visual_language_dataset} with fixed, untuned prompts and exact string matching; model identifiers and decoding and reasoning settings are listed in supplementary Tab.~\ref{tab:vlm_settings}, and pairs, views, prompts, and raw responses are released. A seventh entry, Qwen3-VL~8B LoRA-tuned on Task~1 training pairs (supplementary Sec.~\ref{app:finetune}), uses the same harness and is listed as \emph{Qwen3-VL 8B (FT)}; bold marks compare zero-shot models only.

\paragraph{Task 1: Drawing--Render Verification.}
Models judge whether a rendered view and a 2D drawing depict the same part:
\begin{researchquestion}
\emph{You are shown a rendered view of a CAD model and a 2D engineering drawing. The two images may show the part from different viewpoints. Do they depict the same part? Answer with exactly one word: Yes or No.}
\end{researchquestion}
The drawing is always the front view; the render is a uniformly sampled canonical view of the same part (50\%) or of a different one (50\%), the latter taken from the whole evaluation set (25\%) or from the same surface-type class (25\%). In five of six positive cases the render viewpoint differs from the drawing, so silhouette matching fails even on true positives. Dimensioned and undimensioned drawings are evaluated separately. A control tests whether the forced-choice format suppresses reasoning and, more to the point, whether making the view change explicit in language helps: every model is also run on the undimensioned pairs with a chain-of-thought (CoT) variant that asks it to first think step by step about silhouette, proportions, and distinctive details (holes, slots, steps), and about how the drawing would look from the render's viewpoint, then answer Yes or No on a new final line; only that line is scored, and unparseable responses are excluded and counted.

\paragraph{Task 2: Drawing-to-Render Retrieval.}
Given the right-view drawing of a part, models pick the matching part among four randomly ordered right-view renders, one correct and three distractors sampled uniformly from the evaluation set (chance $=0.25$). Distractor difficulty is not controlled, so some items are easy; the breakdown by surface-type class is in the supplementary. This is a small-gallery analogue of Sec.~\ref{sec:visual_foundational_models}: can a \gls{vlm} do explicitly what frozen features do through embedding similarity?
\begin{researchquestion}
\emph{The first image is a 2D engineering drawing of a CAD model. The following 4 images are rendered CAD models, labeled A, B, C, D. Which rendered model does the drawing represent? Answer with exactly one letter.}
\end{researchquestion}

\paragraph{Task 3: Unguided Complexity Probe.}
Given all six orthographic views, models label the part \emph{Simple} or \emph{Complex}. The prompt deliberately gives no criterion: the task probes whether a \gls{vlm}'s native notion of visual complexity tracks the prismatic/freeform distinction of Sec.~\ref{sec:visual_language_dataset}, whose defining signal (surface type) is only indirectly visible in projection. It is diagnostic rather than a capability measure.
\begin{researchquestion}
\emph{You are shown six views (front, back, left, right, top, bottom) of a CAD model. Classify the model as Simple or Complex based on its geometry. Answer with exactly one word: Simple or Complex.}
\end{researchquestion}

\subsubsection{Empirical Findings.}
\label{sec:vlm_findings}
\paragraph{Verification.}
Table~\ref{tab:main_results} summarizes all tasks. On Task~1 the six zero-shot models score 0.52--0.67 with or without dimensions, Per-model 95\% intervals are about $\pm0.03$ at $n{=}1000$, but since every model answers the same items we compare models with paired exact McNemar tests on the items where their answers differ, Holm-corrected within each family of comparisons (supplementary Tab.~\ref{tab:paired_tests}); no pair among the four leading models differs in either condition ($p_{\text{Holm}}\ge0.28$). Table~\ref{tab:bias_dim} shows why: every model discriminates above chance, but a model-specific decision bias dominates its errors. Claude and GPT-5.6 lean toward \texttt{Yes}; Gemini, Mistral-S, and InternVL3.5 toward \texttt{No}; GPT-5.6~Luna is the most balanced ($|\Delta| = 0.066$). Dimensions push five of the six further toward \texttt{Yes} without a gain in accuracy; the only significant change is a loss, Claude's 4.5 points ($p_{\text{Holm}}=0.020$; supplementary Tab.~\ref{tab:paired_tests}).

Errors also depend on viewpoint: positive accuracy peaks on side renders, which rotate about the vertical axis, and except for Gemini is lowest on top and bottom; Claude falls from 0.98 (right) to 0.46 (top) without dimensions (supplementary). Chain-of-thought does not help (Table~\ref{tab:cot}): under a paired McNemar test of each model's CoT and forced-choice answers on the same items, no model gains significantly (largest gain Claude, $+3.8$ points, $p_{\text{Holm}}=0.17$), the best CoT result (0.681, GPT-5.6~Luna) is $+0.7$ points over its forced-choice answers ($p_{\text{Holm}}=1.0$), and Qwen3-VL loses 4.9 points ($p_{\text{Holm}}=0.007$; supplementary Tab.~\ref{tab:paired_tests}). CoT instead shifts each bias (Claude and GPT-5.6 flip toward rejecting, Mistral-S toward accepting), so the textual intermediate relocates the prior rather than supplying the view relation the task needs. The fine-tuned Qwen3-VL ignores the CoT instruction and reproduces its forced-choice predictions.

\paragraph{Retrieval.}
Gemini leads Task~2 by a wide margin (6.3 points over GPT-5.6~Luna on the items both answered, paired McNemar $p<0.001$), GPT-5.6~Luna is the only other model far above chance, and Claude and Qwen3-VL are only modestly above it (supplementary), yet Gemini is mid-pack on Task~1: relative comparison appears easier than an absolute judgment. Open-source models stay near chance here (InternVL3.5 and Mistral-S are indistinguishable from it, Qwen3-VL is 7 points above) while equalling the closed-source leaders on Task~1, so their gap is specific to choosing among candidates.

\begin{figure}[t]
    \centering
    \includegraphics[width=\linewidth]{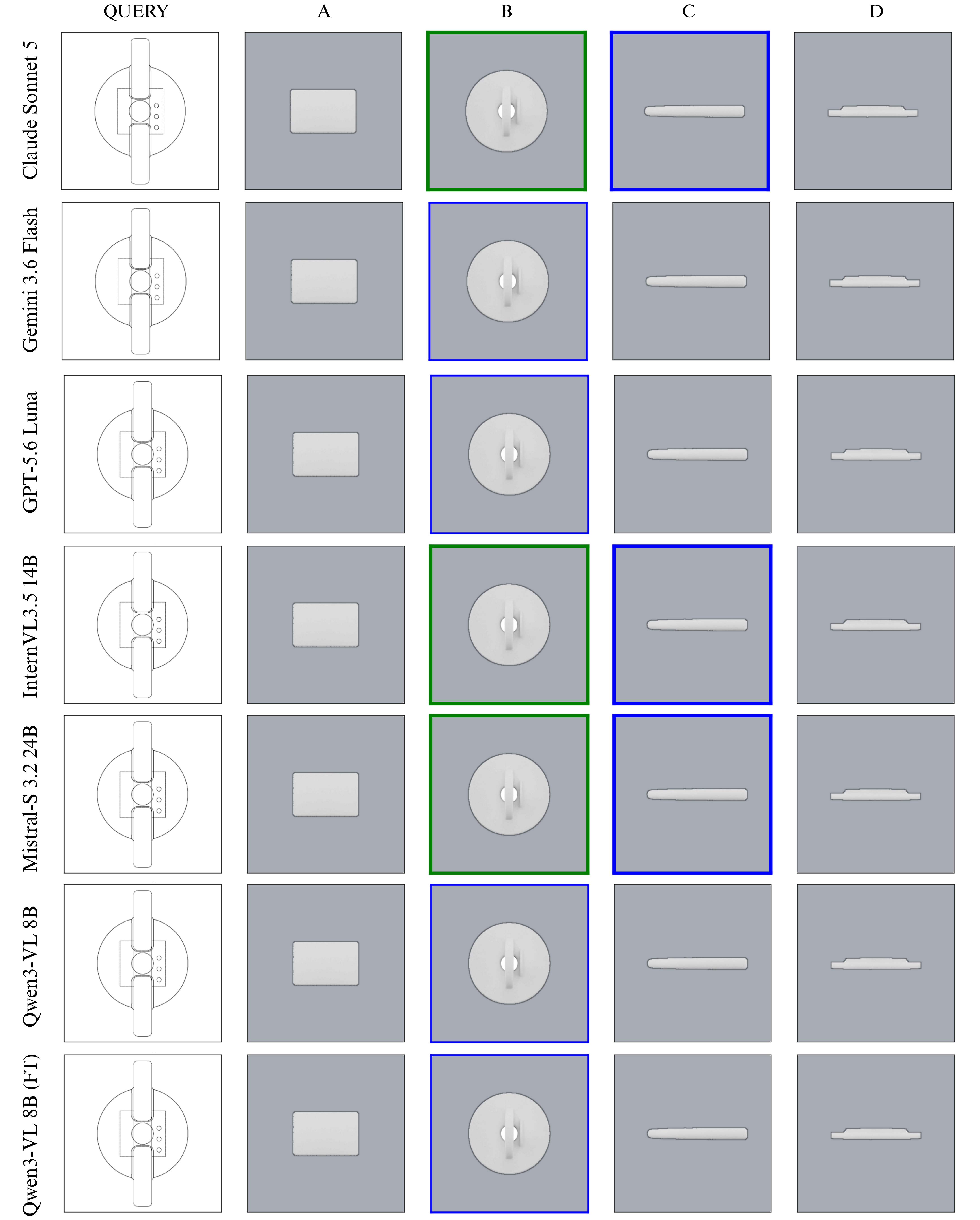}
    \caption{\textbf{Examples of 2D drawing retrieval tasks with four candidate CAD renders.} Green border: ground-truth render; blue border: model selection. A correct selection makes the two borders coincide.}
    \label{fig:task3_samples}
\end{figure}

\paragraph{Complexity probe.}
On Task~3 (Table~\ref{tab:task4_collapse}), every model except GPT-5.6 Luna answers \emph{Simple} on 88--98\% of a balanced set, so near-chance accuracy comes almost entirely from that one label. Asked without a criterion, these models' notion of complexity does not follow surface type: a part dominated by B-spline faces can present as a plain silhouette in projection. GPT-5.6 Luna balances the two per-class recalls, so the signal is recoverable from the same input. In addition, the small Qwen3-VL~8B finetune shows interesting results on this task even without being trained at all on it.

\paragraph{Fine-tuning.}
Rank-16 LoRA adapters on the language model and vision encoder (52.5M parameters, 0.60\%; Table~\ref{tab:ft_config}), trained on 20{,}000 Task~1 pairs, raise Qwen3-VL~8B from 0.670 to 0.685 without dimensions, which is not significant (paired McNemar $p_{\text{Holm}}=0.39$), and from 0.660 to 0.699 with them ($p_{\text{Holm}}=0.031$). The gain comes from negatives: the adapted model rejects 0.930 of negative pairs (zero-shot 0.729) but accepts only 0.440 of positives (0.610), so its lean toward \texttt{No} grows from $-0.119$ to $-0.490$ (Table~\ref{tab:bias_dim}), while top and bottom renders stay at 0.14 and 0.19 on positives. On Task~2 it drops to chance (0.241, 7.6 points below zero-shot, $p_{\text{Holm}}<0.001$); on Task~3 its \emph{Simple} rate falls from 98.4\% to 84.9\% (accuracy 0.512 to 0.553, $p_{\text{Holm}}<0.001$). At this scale, in-domain data buys a few points by shifting the threshold toward \texttt{No}, as its silhouette-mined negatives (70\% of pairs, Sec.~\ref{app:finetune}) predict; whether more data or test-matched negatives change this is open.

\begin{table}[t]
    \centering
    \caption{\textbf{Main VLM results (accuracy, macro-F1)}. Task~1: verification without and with dimensions (DIM), $n{=}1000$ for zero-shot models (one item lacks its render; supplementary Sec.~\ref{app:vlm_settings}) and 1{,}000 for FT; per-model 95\% Wilson intervals are at most $\pm0.031$, and paired model comparisons are in supplementary Tab.~\ref{tab:paired_tests}. Task~2: four-way retrieval (chance 0.25). Task~3: unguided complexity probe scored against surface-type class (chance 0.50). Bold: best model per column, not including the fine-tuned (FT) row.}
    \label{tab:main_results}
    \resizebox{\columnwidth}{!}{%
    \begin{tabular}{lcccccccc}
    \toprule
    \rowcolor{gray!20}
     & \multicolumn{2}{c}{\textbf{Task 1}} & \multicolumn{2}{c}{\textbf{Task 1 + DIM}} & \multicolumn{2}{c}{\textbf{Task 2}} & \multicolumn{2}{c}{\textbf{Task 3}} \\
    \rowcolor{gray!20}
    \textbf{Model} & Acc. $\uparrow$ & F1 $\uparrow$ & Acc. $\uparrow$ & F1 $\uparrow$ & Acc. $\uparrow$ & F1 $\uparrow$ & Acc. $\uparrow$ & F1 $\uparrow$ \\
    \midrule
    \multicolumn{9}{l}{\textit{Closed-source models}} \\
    Claude Sonnet 5   & 0.635 & 0.631 & 0.590 & 0.560 & 0.374 & 0.372 & 0.520 & 0.409 \\
    Gemini 3.6 Flash  & 0.654 & 0.620 & \textbf{0.670} & 0.646 & \textbf{0.690} & \textbf{0.689} & 0.562 & 0.491 \\
    GPT-5.6 Luna      & \textbf{0.670} & \textbf{0.670} & 0.654 & 0.649 & 0.637 & 0.636 & \textbf{0.676} & \textbf{0.676} \\
    \midrule
    \multicolumn{9}{l}{\textit{Open-source models}} \\
    Mistral-S 3.2 24B & 0.534 & 0.417 & 0.524 & 0.450 & 0.274 & 0.235 & 0.521 & 0.409 \\
    InternVL3.5 14B   & 0.624 & 0.604 & 0.637 & 0.618 & 0.266 & 0.274 & 0.506 & 0.399 \\
    Qwen3-VL 8B       & \textbf{0.670} & 0.669 & 0.660 & \textbf{0.655} & 0.317 & 0.305 & 0.512 & 0.363 \\
    \midrule
    Qwen3-VL 8B (FT)  & 0.685 & 0.665 & 0.699 & 0.688 & 0.241 & 0.246 & 0.553 & 0.491 \\
    \bottomrule
    \end{tabular}}
\end{table}

\begin{table}[t]
    \centering
    \caption{\textbf{Complexity probe (Task~3)}: per-class recall, overall accuracy and macro-F1. Labels are B-rep-derived (prismatic = Simple, freeform = Complex); the task probes alignment between an unguided notion of complexity and surface type.}
    \label{tab:task4_collapse}
    \resizebox{\columnwidth}{!}{%
    \begin{tabular}{lcccc}
    \toprule \rowcolor{gray!20}
    \textbf{Model} & \textbf{Acc.} $\uparrow$ & \textbf{Macro F1} $\uparrow$ & \textbf{\texttt{Complex} Rec.} $\uparrow$ & \textbf{\texttt{Simple} Rec.} $\uparrow$ \\
    \midrule
    \multicolumn{5}{l}{\textit{Closed-source models}} \\
    Claude Sonnet 5   & 0.520 & 0.409 & 0.086 & 0.954 \\
    Gemini 3.6 Flash  & 0.562 & 0.491 & 0.188 & 0.940 \\
    GPT-5.6 Luna      & \textbf{0.676} & \textbf{0.676} & 0.712 & 0.640 \\
    \midrule
    \multicolumn{5}{l}{\textit{Open-source models}} \\
    Mistral-S 3.2 24B & 0.521 & 0.409 & 0.087 & 0.954 \\
    InternVL3.5 14B   & 0.506 & 0.399 & 0.084 & 0.928 \\
    Qwen3-VL 8B       & 0.512 & 0.363 & 0.028 & 0.996 \\
    \midrule
    Qwen3-VL 8B (FT)  & 0.553 & 0.491 & 0.204 & 0.902 \\
    \bottomrule
    \end{tabular}}
\end{table}

\begin{table}[t]
    \centering
    \caption{\textbf{Task~1 response bias:} accuracy on positive ($n{=}500$) and negative ($n{=}499$; 500 for FT) pairs, without and with dimensions. $\Delta$ = Pos.\ $-$ Neg.; $\Delta_{\text{DIM}}$ = change in Pos.\ from adding dimensions. Bold: best zero-shot model per column and smallest $|\Delta|$.}
    \label{tab:bias_dim}
    \resizebox{\columnwidth}{!}{%
    \begin{tabular}{lccccccc}
    \toprule
    \rowcolor{gray!20}
     & \multicolumn{3}{c}{\textbf{Undimensioned}} & \multicolumn{3}{c}{\textbf{Dimensioned}} & \\
    \cmidrule(lr){2-4}\cmidrule(lr){5-7}
    \rowcolor{gray!10}
    \textbf{Model} & Pos. $\uparrow$ & Neg. $\uparrow$ & $\Delta$ & Pos. $\uparrow$ & Neg. $\uparrow$ & $\Delta$ & $\Delta_{\text{DIM}}$ \\
    \midrule
    \multicolumn{8}{l}{\textit{Closed-source models}} \\
    Claude Sonnet 5   & \textbf{0.728} & 0.541 & $+0.187$ & \textbf{0.848} & 0.331 & $+0.517$ & $+0.120$ \\
    Gemini 3.6 Flash  & 0.356 & 0.952 & $-0.596$ & 0.412 & \textbf{0.928} & $-0.516$ & $+0.056$ \\
    GPT-5.6 Luna      & 0.703 & 0.637 & $\mathbf{+0.066}$ & 0.772 & 0.535 & $\mathbf{+0.237}$ & $+0.069$ \\
    \midrule
    \multicolumn{8}{l}{\textit{Open-source models}} \\
    Mistral-S 3.2 24B & 0.086 & \textbf{0.982} & $-0.896$ & 0.158 & 0.890 & $-0.732$ & $+0.072$ \\
    InternVL3.5 14B   & 0.400 & 0.848 & $-0.448$ & 0.416 & 0.858 & $-0.442$ & $+0.016$ \\
    Qwen3-VL 8B       & 0.610 & 0.729 & $-0.119$ & 0.540 & 0.780 & $-0.240$ & $-0.070$ \\
    \midrule
    Qwen3-VL 8B (FT)  & 0.440 & 0.930 & $-0.490$ & 0.510 & 0.888 & $-0.378$ & $+0.070$ \\
    \bottomrule
    \end{tabular}}
\end{table}

\begin{table}[t]
    \centering
    \caption{\textbf{Task~1 under chain-of-thought prompting (undimensioned).} Accuracy, macro-F1, positive- and negative-pair accuracy, and $\Delta$ = Pos.\ $-$ Neg., with the forced-choice $\Delta$ from Table~\ref{tab:bias_dim} in parentheses. $n$: items with a parseable final \texttt{Yes}/\texttt{No}. Bold: best zero-shot model per column and smallest $|\Delta|$.}
    \label{tab:cot}
    \resizebox{\columnwidth}{!}{%
    \begin{tabular}{lcccccc}
    \toprule
    \rowcolor{gray!20}
    \textbf{Model} & Acc. $\uparrow$ & F1 $\uparrow$ & Pos. $\uparrow$ & Neg. $\uparrow$ & $\Delta$ (forced-choice) & $n$ \\
    \midrule
    \multicolumn{7}{l}{\textit{Closed-source models}} \\
    Claude Sonnet 5   & 0.673 & 0.664 & 0.515 & 0.830 & $-0.315$ ($+0.187$) & 996 \\
    Gemini 3.6 Flash  & 0.667 & 0.639 & 0.386 & \textbf{0.949} & $-0.563$ ($-0.596$) & 989 \\
    GPT-5.6 Luna      & \textbf{0.681} & \textbf{0.679} & 0.604 & 0.759 & $\mathbf{-0.155}$ ($+0.066$) & 959 \\
    \midrule
    \multicolumn{7}{l}{\textit{Open-source models}} \\
    Mistral-S 3.2 24B & 0.545 & 0.523 & \textbf{0.758} & 0.331 & $+0.427$ ($-0.896$) & 1000 \\
    InternVL3.5 14B   & 0.617 & 0.577 & 0.312 & 0.922 & $-0.610$ ($-0.448$) & 1000 \\
    Qwen3-VL 8B       & 0.621 & 0.610 & 0.452 & 0.790 & $-0.338$ ($-0.119$) & 1000 \\
    \midrule
    Qwen3-VL 8B (FT)  & 0.685 & 0.665 & 0.440 & 0.930 & $-0.490$ ($-0.490$) & 1000 \\
    \bottomrule
    \end{tabular}}
\end{table}
\section{Discussion}
The two paradigms fail in different ways. Retrieval is a relative task, and frozen encoders handle it once view correspondence is given: objectives as different as self-distillation, image--text contrast, and multi-teacher distillation converge, and matching six outlines already comes within 11--17 R@1 points of them. Their advantage grows without viewpoints, where they retrieve about twice as often, so what pretraining contributes is a description of the part that stays consistent across views. \gls{vlm}s face the harder absolute task of verifying a single pair, and their errors are systematic: each applies its own \emph{Yes}/\emph{No} prior, reasoning prompts move that prior without correcting it, a basic fine-tune sharpens it, and top and bottom renders defeat every model, so the bottleneck is the relation between views rather than the detail within them. Handing an encoder the view correspondence is worth more than any change of backbone, while asking a \gls{vlm} to spell it out in words only moves its decision threshold; the task appears to need an intermediate state that keeps the part's geometry and can be re-projected, which neither paradigm carries and the released view labels make testable.

\section{Scope and Limitations}
\label{sec:limitations}
Blueprint2CAD compares line drawings with 2D renders and makes no claim about 3D CAD understanding. Its drawings are clean orthographic projections from a single procedural generator, without section views, GD\&T, title blocks, or scan artifacts; the released pipeline rebuilds the benchmark from other STEP collections, and real documents are the next test. View-aligned scoring uses oracle viewpoint labels and is a diagnostic beside the deployable view-agnostic setting. Evaluation uses frozen weights or zero-shot prompting plus one fine-tuned variant, and the prismatic/freeform classes follow B-rep surface type rather than expert judgments of complexity. Section views, complex drawing collections, and larger-scale fine-tuning are the natural next steps for this benchmark. We based this analysis completely on \cite{ataei2026zerotocadagenticsynthesisinterpretable} dataset, future work could focus on extending this paper to other 3D CAD datasets such as \cite{kou2026mechvqabenchmarkingenhancingmultimodal, koch2019abc, willis2021fusion360} that could have different characteristics.

\section{Conclusion}
Blueprint2CAD makes drawing-to-render correspondence measurable, and BlueCAD-30K supplies the aligned data it needs. Frozen encoders retrieve parts at gallery scale once view correspondence is given, largely independent of pretraining objective, and unknown viewpoints cost far more than dimension lines. VLMs verify pairs through a label prior that chain-of-thought shifts but does not remove, and fail on top and bottom renders even when they discriminate well at a fixed view; LoRA fine-tuning leaves both in place. The bottleneck is neither resolution nor backbone but the relation between views, and whether a geometry-preserving textual intermediate closes it remains open. Section views, larger drawing collections, and larger-scale fine-tuning are natural next steps, and the released pairs and view labels make each testable.

\clearpage
{
    \small
    \bibliographystyle{ieeenat_fullname}
    \bibliography{main}

@String(CVPR= {IEEE Conf. Comput. Vis. Pattern Recog.})

@String(ICCV= {Int. Conf. Comput. Vis.})

@String(ECCV= {Eur. Conf. Comput. Vis.})

@String(ICLR = {Int. Conf. Learn. Represent.})

@String(AAAI = {AAAI})

@String(CVPR  = {CVPR})

@String(ICCV  = {ICCV})

@String(ECCV  = {ECCV})

@String(ICLR  = {ICLR})

@InProceedings{Avetisyan_2019_CVPR,
author = {Avetisyan, Armen and Dahnert, Manuel and Dai, Angela and Savva, Manolis and Chang, Angel X. and Niessner, Matthias},
title = {Scan2CAD: Learning CAD Model Alignment in RGB-D Scans},
booktitle = {The IEEE Conference on Computer Vision and Pattern Recognition (CVPR)},
month = {June},
year = {2019}
}

@inproceedings{han2020spare3d,
  title     = {{SPARE3D}: A Dataset for {SPA}tial {RE}asoning on Three-View Line Drawings},
  author    = {Han, Wenyu and Xiang, Siyuan and Liu, Chenhui and Wang, Ruoyu and Feng, Chen},
  booktitle = {Proceedings of the IEEE/CVF Conference on Computer Vision and Pattern Recognition (CVPR)},
  year      = {2020}
}

@inproceedings{kuo2020mask2cad,
  title     = {{Mask2CAD}: {3D} Shape Prediction by Learning to Segment and Retrieve},
  author    = {Kuo, Weicheng and Angelova, Anelia and Lin, Tsung-Yi and Dai, Angela},
  booktitle = {Proceedings of the European Conference on Computer Vision (ECCV)},
  year      = {2020}
}

@inproceedings{kuo2021patch2cad,
  title     = {{Patch2CAD}: Patchwise Embedding Learning for In-the-Wild Shape Retrieval from a Single Image},
  author    = {Kuo, Weicheng and Angelova, Anelia and Lin, Tsung-Yi and Dai, Angela},
  booktitle = {Proceedings of the IEEE/CVF International Conference on Computer Vision (ICCV)},
  year      = {2021}
}

@inproceedings{gumeli2022roca,
  title     = {{ROCA}: Robust {CAD} Model Retrieval and Alignment from a Single Image},
  author    = {G{\"u}meli, Can and Dai, Angela and Nie{\ss}ner, Matthias},
  booktitle = {Proceedings of the IEEE/CVF Conference on Computer Vision and Pattern Recognition (CVPR)},
  year      = {2022}
}

@article{gao2024diffcad,
  title     = {{DiffCAD}: Weakly-Supervised Probabilistic {CAD} Model Retrieval and Alignment from an {RGB} Image},
  author    = {Gao, Daoyi and Rozenberszki, D{\'a}vid and Leutenegger, Stefan and Dai, Angela},
  journal   = {ACM Transactions on Graphics (SIGGRAPH)},
  volume    = {43},
  number    = {4},
  year      = {2024}
}

@inproceedings{maninis2023cadestate,
  title     = {{CAD-Estate}: Large-Scale {CAD} Model Annotation in {RGB} Videos},
  author    = {Maninis, Kevis-Kokitsi and Popov, Stefan and Nie{\ss}ner, Matthias and Ferrari, Vittorio},
  booktitle = {Proceedings of the IEEE/CVF International Conference on Computer Vision (ICCV)},
  year      = {2023}
}

@inproceedings{dai2017correlated,
  title     = {Deep Correlated Metric Learning for Sketch-Based {3D} Shape Retrieval},
  author    = {Dai, Guoxian and Xie, Jin and Zhu, Fan and Fang, Yi},
  booktitle = {Proceedings of the AAAI Conference on Artificial Intelligence},
  year      = {2017}
}

@inproceedings{chen2018sketch3d,
  title     = {Deep Cross-Modality Adaptation via Semantics Preserving Adversarial Learning for Sketch-Based {3D} Shape Retrieval},
  author    = {Chen, Jiaxin and Fang, Yi},
  booktitle = {Proceedings of the European Conference on Computer Vision (ECCV)},
  year      = {2018}
}

@article{lei2019p2s,
  title     = {Deep Point-to-Subspace Metric Learning for Sketch-Based {3D} Shape Retrieval},
  author    = {Lei, Yinjie and Zhou, Ziqin and Zhang, Pingping and Guo, Yulan and Ma, Zijun and Liu, Lingqiao},
  journal   = {Pattern Recognition},
  volume    = {96},
  year      = {2019}
}

@inproceedings{li2013shrec,
  title     = {{SHREC}'13 Track: Large Scale Sketch-Based {3D} Shape Retrieval},
  author    = {Li, Bo and Lu, Yijuan and Godil, Afzal and Schreck, Tobias and Aono, Masaki and Johan, Henry and Saavedra, Jose M. and Tashiro, Shoki},
  booktitle = {Eurographics Workshop on 3D Object Retrieval (3DOR)},
  year      = {2013}
}

@article{qin2022shrec,
  title     = {{SHREC}'22 Track: Sketch-Based {3D} Shape Retrieval in the Wild},
  author    = {Qin, Jie and Yuan, Shuaihang and Chen, Jiaxin and others},
  journal   = {Computers \& Graphics},
  volume    = {107},
  pages     = {104--115},
  year      = {2022}
}

@article{manda2021cadsketchnet,
  title     = {{CADSketchNet} -- An Annotated Sketch Dataset for {3D} {CAD} Model Retrieval with Deep Neural Networks},
  author    = {Manda, Bharadwaj and Dhayarkar, Shubham and Mitheran, Sai and Viekash, V. K. and Muthuganapathy, Ramanathan},
  journal   = {Computers \& Graphics},
  volume    = {99},
  pages     = {100--113},
  year      = {2021}
}

@inproceedings{radford2021clip,
  title     = {Learning Transferable Visual Models from Natural Language Supervision},
  author    = {Radford, Alec and Kim, Jong Wook and Hallacy, Chris and others},
  booktitle = {Proceedings of the International Conference on Machine Learning (ICML)},
  year      = {2021}
}

@article{oquab2023dinov2,
  title     = {{DINOv2}: Learning Robust Visual Features without Supervision},
  author    = {Oquab, Maxime and Darcet, Timoth{\'e}e and Moutakanni, Th{\'e}o and others},
  journal   = {Transactions on Machine Learning Research},
  year      = {2024}
}

@misc{simeoni2025dinov3,
      title={DINOv3}, 
      author={Oriane Siméoni and Huy V. Vo and Maximilian Seitzer and Federico Baldassarre and Maxime Oquab and Cijo Jose and Vasil Khalidov and Marc Szafraniec and Seungeun Yi and Michaël Ramamonjisoa and Francisco Massa and Daniel Haziza and Luca Wehrstedt and Jianyuan Wang and Timothée Darcet and Théo Moutakanni and Leonel Sentana and Claire Roberts and Andrea Vedaldi and Jamie Tolan and John Brandt and Camille Couprie and Julien Mairal and Hervé Jégou and Patrick Labatut and Piotr Bojanowski},
      year={2025},
      eprint={2508.10104},
      archivePrefix={arXiv},
      primaryClass={cs.CV},
      url={https://arxiv.org/abs/2508.10104}
}

@inproceedings{kirillov2023sam,
  title     = {Segment Anything},
  author    = {Kirillov, Alexander and Mintun, Eric and Ravi, Nikhila and others},
  booktitle = {Proceedings of the IEEE/CVF International Conference on Computer Vision (ICCV)},
  year      = {2023}
}

@inproceedings{xue2023ulip,
  title     = {{ULIP}: Learning a Unified Representation of Language, Images, and Point Clouds for {3D} Understanding},
  author    = {Xue, Le and Gao, Mingfei and Xing, Chen and Mart{\'i}n-Mart{\'i}n, Roberto and Wu, Jiajun and Xiong, Caiming and Xu, Ran and Niebles, Juan Carlos and Savarese, Silvio},
  booktitle = {Proceedings of the IEEE/CVF Conference on Computer Vision and Pattern Recognition (CVPR)},
  year      = {2023}
}

@inproceedings{dettmers2023qlora,
  title={{QLoRA}: Efficient Finetuning of Quantized {LLMs}},
  author={Dettmers, Tim and Pagnoni, Artidoro and Holtzman, Ari and Zettlemoyer, Luke},
  booktitle={Advances in Neural Information Processing Systems (NeurIPS)},
  year={2023}
}

@misc{hu2021loralowrankadaptationlarge,
      title={LoRA: Low-Rank Adaptation of Large Language Models}, 
      author={Edward J. Hu and Yelong Shen and Phillip Wallis and Zeyuan Allen-Zhu and Yuanzhi Li and Shean Wang and Lu Wang and Weizhu Chen},
      year={2021},
      eprint={2106.09685},
      archivePrefix={arXiv},
      primaryClass={cs.CL},
      url={https://arxiv.org/abs/2106.09685}, 
}

@inproceedings{xue2024ulip2,
  title     = {{ULIP-2}: Towards Scalable Multimodal Pre-Training for {3D} Understanding},
  author    = {Xue, Le and Yu, Ning and Zhang, Shu and others},
  booktitle = {Proceedings of the IEEE/CVF Conference on Computer Vision and Pattern Recognition (CVPR)},
  year      = {2024}
}

@InProceedings{Ranzinger_2024_CVPR,
    author    = {Ranzinger, Mike and Heinrich, Greg and Kautz, Jan and Molchanov, Pavlo},
    title     = {AM-RADIO: Agglomerative Vision Foundation Model Reduce All Domains Into One},
    booktitle = {Proceedings of the IEEE/CVF Conference on Computer Vision and Pattern Recognition (CVPR)},
    month     = {June},
    year      = {2024},
    pages     = {12490-12500}
}

@inproceedings{liu2023openshape,
  title     = {{OpenShape}: Scaling Up {3D} Shape Representation Towards Open-World Understanding},
  author    = {Liu, Minghua and Shi, Ruoxi and Kuang, Kaiming and others},
  booktitle = {Advances in Neural Information Processing Systems (NeurIPS)},
  year      = {2023}
}

@inproceedings{zhou2024uni3d,
  title     = {{Uni3D}: Exploring Unified {3D} Representation at Scale},
  author    = {Zhou, Junsheng and Wang, Jinsheng and Ma, Baorui and Liu, Yu-Shen and Huang, Tiejun and Wang, Xinlong},
  booktitle = {Proceedings of the International Conference on Learning Representations (ICLR)},
  year      = {2024}
}

@article{yuan2024openecad,
  title     = {{OpenECAD}: An Efficient Visual Language Model for Editable {3D-CAD} Design},
  author    = {Yuan, Zhe and Shi, Jianqi and Huang, Yanhong},
  journal   = {Computers \& Graphics},
  volume    = {124},
  pages     = {104048},
  year      = {2024}
}

@article{xu2024cadmllm,
  title     = {{CAD-MLLM}: Unifying Multimodality-Conditioned {CAD} Generation with {MLLM}},
  author    = {Xu, Jingwei and Wang, Chenyu and Zhao, Zibo and Liu, Wen and Ma, Yi and Gao, Shenghua},
  journal   = {arXiv preprint arXiv:2411.04954},
  year      = {2024}
}

@inproceedings{wang2025cadgpt,
  title     = {{CAD-GPT}: Synthesising {CAD} Construction Sequence with Spatial Reasoning-Enhanced Multimodal {LLMs}},
  author    = {Wang, Siyu and Chen, Cailian and Le, Xinyi and Xu, Qimin and Xu, Lei and Zhang, Yanzhou and Chen, Jie},
  booktitle = {Proceedings of the AAAI Conference on Artificial Intelligence},
  year      = {2025}
}

@inproceedings{chen2025cadcrafter,
  title     = {{CADCrafter}: Generating Computer-Aided Design Models from Unconstrained Images},
  author    = {Chen, Cheng and Wei, Jiacheng and Chen, Tianrun and others},
  booktitle = {Proceedings of the IEEE/CVF Conference on Computer Vision and Pattern Recognition (CVPR)},
  year      = {2025}
}

@inproceedings{wu2024cadvlm,
  title     = {{CadVLM}: Bridging Language and Vision in the Generation of Parametric {CAD} Sketches},
  author    = {Wu, Sifan and Khasahmadi, Amir Hosein and Katz, Mor and Jayaraman, Pradeep Kumar and Pu, Yewen and Willis, Karl and Liu, Bang},
  booktitle = {Proceedings of the European Conference on Computer Vision (ECCV)},
  year      = {2024}
}

@inproceedings{rukhovich2025cadrecode,
  title     = {{CAD-Recode}: Reverse Engineering {CAD} Code from Point Clouds},
  author    = {Rukhovich, Danila and Dupont, Elona and Mallis, Dimitrios and Cherenkova, Kseniya and Kacem, Anis and Aouada, Djamila},
  booktitle = {Proceedings of the IEEE/CVF International Conference on Computer Vision (ICCV)},
  year      = {2025}
}

@article{chen2024img2cad,
  title     = {{Img2CAD}: Conditioned {3D} {CAD} Model Generation from Single Image with Structured Visual Geometry},
  author    = {Chen, Tianrun and Yu, Chunan and Hu, Yuanqi and others},
  journal   = {arXiv preprint arXiv:2410.03417},
  year      = {2024}
}

@article{you2024img2cad,
  title     = {{Img2CAD}: Reverse Engineering {3D} {CAD} Models from Images through {VLM}-Assisted Conditional Factorization},
  author    = {You, Yang and Uy, Mikaela Angelina and Han, Jiaqi and others},
  journal   = {arXiv preprint arXiv:2408.01437},
  year      = {2024}
}

@inproceedings{wu2021deepcad,
  title     = {{DeepCAD}: A Deep Generative Network for Computer-Aided Design Models},
  author    = {Wu, Rundi and Xiao, Chang and Zheng, Changxi},
  booktitle = {Proceedings of the IEEE/CVF International Conference on Computer Vision (ICCV)},
  year      = {2021}
}

@inproceedings{xu2022skexgen,
  title     = {{SkexGen}: Autoregressive Generation of {CAD} Construction Sequences with Disentangled Codebooks},
  author    = {Xu, Xiang and Willis, Karl D. D. and Lambourne, Joseph G. and Cheng, Chin-Yi and Jayaraman, Pradeep Kumar and Furukawa, Yasutaka},
  booktitle = {Proceedings of the International Conference on Machine Learning (ICML)},
  year      = {2022}
}

@inproceedings{xu2023hnccad,
  title     = {Hierarchical Neural Coding for Controllable {CAD} Model Generation},
  author    = {Xu, Xiang and Jayaraman, Pradeep Kumar and Lambourne, Joseph G. and Willis, Karl D. D. and Furukawa, Yasutaka},
  booktitle = {Proceedings of the International Conference on Machine Learning (ICML)},
  year      = {2023}
}

@inproceedings{lambourne2021brepnet,
  title     = {{BRepNet}: A Topological Message Passing System for Solid Models},
  author    = {Lambourne, Joseph G. and Willis, Karl D. D. and Jayaraman, Pradeep Kumar and Sanghi, Aditya and Meltzer, Peter and Shayani, Hooman},
  booktitle = {Proceedings of the IEEE/CVF Conference on Computer Vision and Pattern Recognition (CVPR)},
  year      = {2021}
}

@inproceedings{uy2022point2cyl,
  title     = {{Point2Cyl}: Reverse Engineering {3D} Objects from Point Clouds to Extrusion Cylinders},
  author    = {Uy, Mikaela Angelina and Chang, Yen-Yu and Sung, Minhyuk and Goel, Purvi and Lambourne, Joseph G. and Birdal, Tolga and Guibas, Leonidas J.},
  booktitle = {Proceedings of the IEEE/CVF Conference on Computer Vision and Pattern Recognition (CVPR)},
  year      = {2022}
}

@inproceedings{liu2024point2cad,
  title     = {{Point2CAD}: Reverse Engineering {CAD} Models from {3D} Point Clouds},
  author    = {Liu, Yujia and Obukhov, Anton and Wegner, Jan Dirk and Schindler, Konrad},
  booktitle = {Proceedings of the IEEE/CVF Conference on Computer Vision and Pattern Recognition (CVPR)},
  year      = {2024}
}

@inproceedings{khan2024cadsignet,
  title     = {{CAD-SIGNet}: {CAD} Language Inference from Point Clouds using Layer-wise Sketch Instance Guided Attention},
  author    = {Khan, Mohammad Sadil and Dupont, Elona and Ali, Sk Aziz and Cherenkova, Kseniya and Kacem, Anis and Aouada, Djamila},
  booktitle = {Proceedings of the IEEE/CVF Conference on Computer Vision and Pattern Recognition (CVPR)},
  year      = {2024}
}

@inproceedings{khan2024text2cad,
  title     = {{Text2CAD}: Generating Sequential {CAD} Designs from Beginner-to-Expert Level Text Prompts},
  author    = {Khan, Mohammad Sadil and Sinha, Sankalp and Uddin, Sheikh Talha and Stricker, Didier and Ali, Sk Aziz and Afzal, Muhammad Zeshan},
  booktitle = {Advances in Neural Information Processing Systems (NeurIPS)},
  year      = {2024}
}

@article{li2022free2cad,
  title     = {{Free2CAD}: Parsing Freehand Drawings into {CAD} Commands},
  author    = {Li, Changjian and Pan, Hao and Bousseau, Adrien and Mitra, Niloy J.},
  journal   = {ACM Transactions on Graphics (SIGGRAPH)},
  volume    = {41},
  number    = {4},
  year      = {2022}
}

@inproceedings{seff2022vitruvion,
  title     = {{Vitruvion}: A Generative Model of Parametric {CAD} Sketches},
  author    = {Seff, Ari and Zhou, Wenda and Richardson, Nick and Adams, Ryan P.},
  booktitle = {Proceedings of the International Conference on Learning Representations (ICLR)},
  year      = {2022}
}

@inproceedings{hu2023plankassembly,
  title     = {{PlankAssembly}: Robust {3D} Reconstruction from Three Orthographic Views with Learnt Shape Programs},
  author    = {Hu, Wentao and Zheng, Jia and Zhang, Zixin and Yuan, Xiaojun and Yin, Jian and Zhou, Zihan},
  booktitle = {Proceedings of the IEEE/CVF International Conference on Computer Vision (ICCV)},
  year      = {2023}
}

@article{wang2024cad2program,
  title     = {From {2D} {CAD} Drawings to {3D} Parametric Models: A Vision-Language Approach},
  author    = {Wang, Xilin and Zheng, Jia and Hu, Yuanchao and Zhu, Hao and Yu, Qian and Zhou, Zihan},
  journal   = {arXiv preprint arXiv:2412.11892},
  year      = {2024}
}

@inproceedings{qin2025drawing2cad,
  title     = {{Drawing2CAD}: Sequence-to-Sequence Learning for {CAD} Generation from Vector Drawings},
  author    = {Qin, Feiwei and Hou, Ziyan and others},
  booktitle = {Proceedings of the ACM International Conference on Multimedia (ACM MM)},
  year      = {2025}
}

@article{willis2021fusion360,
  title     = {{Fusion 360 Gallery}: A Dataset and Environment for Programmatic {CAD} Construction from Human Design Sequences},
  author    = {Willis, Karl D. D. and Pu, Yewen and Luo, Jieliang and Chu, Hang and Du, Tao and Lambourne, Joseph G. and Solar-Lezama, Armando and Matusik, Wojciech},
  journal   = {ACM Transactions on Graphics (SIGGRAPH)},
  volume    = {40},
  number    = {4},
  year      = {2021}
}

@inproceedings{kim2020mcb,
  title     = {A Large-Scale Annotated Mechanical Components Benchmark for Classification and Retrieval Tasks with Deep Neural Networks},
  author    = {Kim, Sangpil and Chi, Hyung-gun and Hu, Xiao and Huang, Qixing and Ramani, Karthik},
  booktitle = {Proceedings of the European Conference on Computer Vision (ECCV)},
  year      = {2020}
}

@article{seff2020sketchgraphs,
  title     = {{SketchGraphs}: A Large-Scale Dataset for Modeling Relational Geometry in Computer-Aided Design},
  author    = {Seff, Ari and Ovadia, Yaniv and Zhou, Wenda and Adams, Ryan P.},
  journal   = {arXiv preprint arXiv:2007.08506},
  year      = {2020}
}

@misc{ataei2026zerotocadagenticsynthesisinterpretable,
  title={Zero-to-CAD: Agentic Synthesis of Interpretable CAD Programs at Million-Scale Without Real Data}, 
  author={Mohammadmehdi Ataei and Farzaneh Askari and Kamal Rahimi Malekshan and Pradeep Kumar Jayaraman},
  year={2026},
  eprint={2604.24479},
  archivePrefix={arXiv},
  primaryClass={cs.CV},
  url={https://arxiv.org/abs/2604.24479}
}

@inproceedings{sutherland1963sketchpad,
  title     = {Sketchpad: A Man-Machine Graphical Communication System},
  author    = {Sutherland, Ivan E.},
  booktitle = {Proceedings of the AFIPS Spring Joint Computer Conference},
  pages     = {329--346},
  year      = {1963}
}

@article{requicha1980representations,
  title     = {Representations for Rigid Solids: Theory, Methods, and Systems},
  author    = {Requicha, Aristides A. G.},
  journal   = {ACM Computing Surveys},
  volume    = {12},
  number    = {4},
  pages     = {437--464},
  year      = {1980}
}

@article{braid1975synthesis,
  title     = {The Synthesis of Solids Bounded by Many Faces},
  author    = {Braid, Ian C.},
  journal   = {Communications of the ACM},
  volume    = {18},
  number    = {4},
  pages     = {209--216},
  year      = {1975}
}

@book{mantyla1988solid,
  title     = {An Introduction to Solid Modeling},
  author    = {M{\"a}ntyl{\"a}, Martti},
  publisher = {Computer Science Press},
  year      = {1988}
}

@book{piegl1997nurbs,
  title     = {The {NURBS} Book},
  author    = {Piegl, Les and Tiller, Wayne},
  edition   = {2nd},
  publisher = {Springer},
  year      = {1997}
}

@book{shah1995parametric,
  title     = {Parametric and Feature-Based {CAD/CAM}: Concepts, Techniques, and Applications},
  author    = {Shah, Jami J. and M{\"a}ntyl{\"a}, Martti},
  publisher = {Wiley},
  year      = {1995}
}

@book{monge1799geometrie,
  title     = {G{\'e}om{\'e}trie Descriptive},
  author    = {Monge, Gaspard},
  publisher = {Baudouin},
  address   = {Paris},
  year      = {1799}
}

@misc{iso128,
  title        = {{ISO} 128: Technical Product Documentation ({TPD}) --- General Principles of Representation},
  author       = {{International Organization for Standardization}},
  howpublished = {ISO Standard},
  year         = {2020}
}

@misc{autodesk_dxf,
  title        = {{DXF} Reference},
  author       = {{Autodesk, Inc.}},
  howpublished = {Autodesk Developer Documentation},
  year         = {2011}
}

@misc{w3c_svg,
  title        = {Scalable Vector Graphics ({SVG}) 1.1 (Second Edition)},
  author       = {{World Wide Web Consortium}},
  howpublished = {W3C Recommendation},
  year         = {2011}
}

@article{idesawa1973system,
  title     = {A System to Generate a Solid Figure from Three View},
  author    = {Idesawa, Masanori},
  journal   = {Bulletin of the JSME},
  volume    = {16},
  number    = {92},
  pages     = {216--225},
  year      = {1973}
}

@book{giesecke2016technical,
  title     = {Technical Drawing with Engineering Graphics},
  author    = {Giesecke, Frederick E. and Mitchell, Alva and Spencer, Henry C. and Hill, Ivan L. and Dygdon, John T. and Novak, James E. and Lockhart, Shawna and Johnson, Cindy and Goodman, Marla},
  edition   = {15th},
  publisher = {Pearson},
  year      = {2016}
}

@inproceedings{cao2026tipsv2,
  title     = {{TIPSv2: Advancing Vision-Language Pretraining with Enhanced Patch-Text Alignment}},
  author    = {Cao, Bingyi and Chen, Koert and Maninis, Kevis-Kokitsi and Chen, Kaifeng and Karpur, Arjun and Xia, Ye and Dua, Sahil and Dabral, Tanmaya and Han, Guangxing and Han, Bohyung and Ainslie, Joshua and Bewley, Alex and Jacob, Mithun and Wagner, Rene and Ramos, Washington and Choromanski, Krzysztof and Seyedhosseini, Mojtaba and Zhou, Howard and Araujo, Andre},
  booktitle = {Proceedings of the IEEE/CVF Conference on Computer Vision and Pattern Recognition (CVPR)},
  year      = {2026},
  url       = {https://arxiv.org/abs/2604.12012}
}

@article{kolodiazhnyi2025cadrille,
  title   = {cadrille: Multi-modal {CAD} Reconstruction with Online Reinforcement Learning},
  author  = {Kolodiazhnyi, Maksim and Tarasov, Denis and Zhemchuzhnikov, Dmitrii and Nikulin, Alexander and Zisman, Ilya and Vorontsova, Anna and Konushin, Anton and Kurenkov, Vladislav and Rukhovich, Danila},
  journal = {arXiv preprint arXiv:2505.22914},
  year    = {2025}
}

@misc{doris2025cadcoderopensourcevisionlanguagemodel,
      title={CAD-Coder: An Open-Source Vision-Language Model for Computer-Aided Design Code Generation}, 
      author={Anna C. Doris and Md Ferdous Alam and Amin Heyrani Nobari and Faez Ahmed},
      year={2025},
      eprint={2505.14646},
      archivePrefix={arXiv},
      primaryClass={cs.CV},
      url={https://arxiv.org/abs/2505.14646}, 
}

@misc{joglekar2026ortho2cad3dcadgeneration,
      title={Ortho2CAD: 3D CAD generation from orthographic drawings using vision language models}, 
      author={Aditya Joglekar and Amit Regmi and Kenji Shimada and Levent Burak Kara},
      year={2026},
      eprint={2607.08891},
      archivePrefix={arXiv},
      primaryClass={cs.CE},
      url={https://arxiv.org/abs/2607.08891}, 
}

@misc{kou2026mechvqabenchmarkingenhancingmultimodal,
      title={MechVQA: Benchmarking and Enhancing Multimodal LLMs on Comprehensive Mechanical Drawing Understanding}, 
      author={Qian Kou and Xiaofeng Shi and Yulin Li and Xiaosong Qiu and Xinyang Wang and Hua Zhou and Cao Dongxing},
      year={2026},
      eprint={2605.30794},
      archivePrefix={arXiv},
      primaryClass={cs.CV},
      url={https://arxiv.org/abs/2605.30794}, 
}

@misc{niu2025creftcadboostingorthographicprojection,
      title={CReFT-CAD: Boosting Orthographic Projection Reasoning for CAD via Reinforcement Fine-Tuning}, 
      author={Ke Niu and Zhuofan Chen and Haiyang Yu and Yuwen Chen and Teng Fu and Mengyang Zhao and Bin Li and Xiangyang Xue},
      year={2025},
      eprint={2506.00568},
      archivePrefix={arXiv},
      primaryClass={cs.CV},
      url={https://arxiv.org/abs/2506.00568}, 
}

@misc{zhan20263viewsensespatialmentalperspective,
      title={3ViewSense: Spatial and Mental Perspective Reasoning from Orthographic Views in Vision-Language Models}, 
      author={Shaoxiong Zhan and Yanlin Lai and Zheng Liu and Hai Lin and Shen Li and Xiaodong Cai and Zijian Lin and Wen Huang and Hai-Tao Zheng},
      year={2026},
      eprint={2603.07751},
      archivePrefix={arXiv},
      primaryClass={cs.CV},
      url={https://arxiv.org/abs/2603.07751}, 
}

@misc{michalkiewicz2026giqbenchmarking3dgeometric,
      title={GIQ: Benchmarking 3D Geometric Reasoning of Vision Foundation Models with Simulated and Real Polyhedra}, 
      author={Mateusz Michalkiewicz and Anekha Sokhal and Tadeusz Michalkiewicz and Piotr Pawlikowski and Mahsa Baktashmotlagh and Varun Jampani and Guha Balakrishnan},
      year={2026},
      eprint={2506.08194},
      archivePrefix={arXiv},
      primaryClass={cs.CV},
      url={https://arxiv.org/abs/2506.08194}, 
}

@inproceedings{koch2019abc,
  author    = {Koch, Sebastian and Matveev, Albert and Jiang, Zhongshi and
               Williams, Francis and Artemov, Alexey and Burnaev, Evgeny and
               Alexa, Marc and Zorin, Denis and Panozzo, Daniele},
  title     = {{ABC}: A Big {CAD} Model Dataset for Geometric Deep Learning},
  booktitle = {Proceedings of the IEEE/CVF Conference on Computer Vision and
               Pattern Recognition (CVPR)},
  pages     = {9601--9611},
  year      = {2019}
}

@article{arslan2021shrec,
  author    = {Arslan, Mazlum Ferhat and Haridis, Alexandros and
               Rosin, Paul L. and Tari, Sibel and Brassey, Charlotte and
               Gardiner, James D. and Genctav, Asli and Genctav, Murat},
  title     = {{SHREC}'21: Quantifying Shape Complexity},
  journal   = {Computers \& Graphics},
  volume    = {102},
  pages     = {144--153},
  year      = {2022}
}

@inproceedings{jayaraman2021uvnet,
  author    = {Jayaraman, Pradeep Kumar and Sanghi, Aditya and
               Lambourne, Joseph G. and Willis, Karl D. D. and
               Davies, Thomas and Shayani, Hooman and Morris, Nigel},
  title     = {{UV-Net}: Learning from Boundary Representations},
  booktitle = {Proceedings of the IEEE/CVF Conference on Computer Vision and
               Pattern Recognition (CVPR)},
  pages     = {11703--11712},
  year      = {2021}
}

@article{wadell1935volume,
  author    = {Wadell, Hakon},
  title     = {Volume, Shape, and Roundness of Quartz Particles},
  journal   = {The Journal of Geology},
  volume    = {43},
  number    = {3},
  pages     = {250--280},
  year      = {1935}
}

@misc{tschannen2025siglip2multilingualvisionlanguage,
      title={SigLIP 2: Multilingual Vision-Language Encoders with Improved Semantic Understanding, Localization, and Dense Features}, 
      author={Michael Tschannen and Alexey Gritsenko and Xiao Wang and Muhammad Ferjad Naeem and Ibrahim Alabdulmohsin and Nikhil Parthasarathy and Talfan Evans and Lucas Beyer and Ye Xia and Basil Mustafa and Olivier Hénaff and Jeremiah Harmsen and Andreas Steiner and Xiaohua Zhai},
      year={2025},
      eprint={2502.14786},
      archivePrefix={arXiv},
      primaryClass={cs.CV},
      url={https://arxiv.org/abs/2502.14786}, 
}

@misc{mistralsmall2025,
  author       = {{Mistral AI Team}},
  title        = {Mistral Small 3.2 (2506): Optimized 24B Open-Weight Language Model},
  howpublished = {\url{https://docs.mistral.ai/models/mistral-small-3-2-25-06}},
  year         = {2025},
  note         = {Accessed: 2026-08-19}
}

@misc{wang2025internvl35advancingopensourcemultimodal,
      title={InternVL3.5: Advancing Open-Source Multimodal Models in Versatility, Reasoning, and Efficiency}, 
      author={Weiyun Wang and others},
      year={2025},
      eprint={2508.18265},
      archivePrefix={arXiv},
      primaryClass={cs.CV},
      url={https://arxiv.org/abs/2508.18265}, 
}

@misc{bai2025qwen3vltechnicalreport,
      title={Qwen3-VL Technical Report}, 
      author={Shuai Bai and Yuxuan Cai and Ruizhe Chen and Keqin Chen and Xionghui Chen and Zesen Cheng and Lianghao Deng and Wei Ding and Chang Gao and Chunjiang Ge and Wenbin Ge and Zhifang Guo and Qidong Huang and Jie Huang and Fei Huang and Binyuan Hui and Shutong Jiang and Zhaohai Li and Mingsheng Li and Mei Li and Kaixin Li and Zicheng Lin and Junyang Lin and Xuejing Liu and Jiawei Liu and Chenglong Liu and Yang Liu and Dayiheng Liu and Shixuan Liu and Dunjie Lu and Ruilin Luo and Chenxu Lv and Rui Men and Lingchen Meng and Xuancheng Ren and Xingzhang Ren and Sibo Song and Yuchong Sun and Jun Tang and Jianhong Tu and Jianqiang Wan and Peng Wang and Pengfei Wang and Qiuyue Wang and Yuxuan Wang and Tianbao Xie and Yiheng Xu and Haiyang Xu and Jin Xu and Zhibo Yang and Mingkun Yang and Jianxin Yang and An Yang and Bowen Yu and Fei Zhang and Hang Zhang and Xi Zhang and Bo Zheng and Humen Zhong and Jingren Zhou and Fan Zhou and Jing Zhou and Yuanzhi Zhu and Ke Zhu},
      year={2025},
      eprint={2511.21631},
      archivePrefix={arXiv},
      primaryClass={cs.CV},
      url={https://arxiv.org/abs/2511.21631}, 
}

@misc{gemini36flash,
  author={{Google DeepMind}},
  title={Gemini 3.6 Flash},
  year={2026},
  howpublished={Google DeepMind Model Card},
  url={https://deepmind.google/models/gemini/flash/}
}

@misc{openai2026gpt56,
  author={{OpenAI}},
  title={GPT-5.6: Frontier Intelligence That Scales With Your Ambition},
  year={2026},
  howpublished={OpenAI},
  url={https://openai.com/index/gpt-5-6/}
}

@misc{anthropic2026sonnet5,
  author={{Anthropic}},
  title={Introducing Claude Sonnet 5},
  year={2026},
  howpublished={Anthropic},
  url={https://www.anthropic.com/news/claude-sonnet-5}
}

@inproceedings{alrashedy2025cadcodeverify,
  title={Generating CAD Code with Vision-Language Models for 3D Designs},
  author={Alrashedy, Kamel and others},
  booktitle={International Conference on Learning Representations (ICLR)},
  year={2025}
}

@article{dimitrov2023engineering,
  title={Engineering Drawing Understanding: A Survey and Framework for 3D Reconstruction},
  author={Dimitrov, Martin and Fisher, Robert B and Gao, Lin},
  journal={Computer-Aided Design},
  volume={161},
  pages={103524},
  year={2023},
  publisher={Elsevier}
}

@article{moreno2021computer,
  title={Computer vision techniques for automatic detection and recognition of engineering drawing annotations: A review},
  author={Moreno-Garcia, Carlos Francisco and Elyan, Eyad and Jayne, Chrisina},
  journal={Neural Computing and Applications},
  volume={33},
  number={16},
  pages={9843--9864},
  year={2021},
  publisher={Springer}
}

@article{de2022symbol,
  title={Symbol detection and text recognition in architectural and engineering drawings using deep learning},
  author={de las Heras, Lluis-Pau and Terrades, Oriol Ramos and Valveny, Ernest and S{\'a}nchez, Gemma},
  journal={Pattern Recognition Letters},
  volume={156},
  pages={112--118},
  year={2022},
  publisher={Elsevier}
}

@inproceedings{hou2024sketch,
  author    = {Hou, Wentao and Diao, Zhenyu and Peng, Jingliang},
  title     = {Sketch-Based 3D Shape Retrieval via Cross-Modal Contrastive Learning and Difficulty-Aware Uncertainty Regularization},
  booktitle = {Pattern Recognition and Computer Vision (PRCV)},
  series    = {Lecture Notes in Computer Science},
  volume    = {15036},
  pages     = {521--534},
  publisher = {Springer},
  year      = {2024},
  doi       = {10.1007/978-981-97-8508-7_36}
}

@article{su2025thinkingwithimages,
  title     = {Thinking with Images for Multimodal Reasoning: Foundations, Methods, and Future Frontiers},
  author    = {Su, Zhaochen and Xia, Peng and Guo, Hangyu and Liu, Zhenhua and Ma, Yan and Qu, Xiaoye and Liu, Jiaqi and Li, Yanshu and Zeng, Kaide and Yang, Zhengyuan and Li, Linjie and Cheng, Yu and Ji, Heng and He, Junxian and Fung, Yi R.},
  journal   = {arXiv preprint arXiv:2506.23918},
  year      = {2025}
}

@inproceedings{hu2024visualsketchpad,
  title     = {Visual Sketchpad: Sketching as a Visual Chain of Thought for Multimodal Language Models},
  author    = {Hu, Yushi and Shi, Weijia and Fu, Xingyu and Roth, Dan and Ostendorf, Mari and Zettlemoyer, Luke and Smith, Noah A. and Krishna, Ranjay},
  booktitle = {Advances in Neural Information Processing Systems (NeurIPS)},
  year      = {2024}
}

@inproceedings{li2025mvot,
  title     = {Imagine while Reasoning in Space: Multimodal Visualization-of-Thought},
  author    = {Li, Chengzu and Wu, Wenshan and Zhang, Huanyu and Xia, Yan and Mao, Shaoguang and Dong, Li and Vuli{\'c}, Ivan and Wei, Furu},
  booktitle = {Proceedings of the International Conference on Machine Learning (ICML)},
  year      = {2025}
}

@inproceedings{bigverdi2025perceptiontokens,
  title     = {Perception Tokens Enhance Visual Reasoning in Multimodal Language Models},
  author    = {Bigverdi, Mahtab and Luo, Zelun and Hsieh, Cheng-Yu and Shen, Ethan and Chen, Dongping and Shapiro, Linda G. and Krishna, Ranjay},
  booktitle = {Proceedings of the IEEE/CVF Conference on Computer Vision and Pattern Recognition (CVPR)},
  year      = {2025}
}

@inproceedings{yang2025mirage,
  title     = {Machine Mental Imagery: Empower Multimodal Reasoning with Latent Visual Tokens},
  author    = {Yang, Zeyuan and Yu, Xueyang and Chen, Delin and Shen, Maohao and Gan, Chuang},
  booktitle = {Proceedings of the IEEE/CVF Conference on Computer Vision and Pattern Recognition (CVPR)},
  year      = {2026}
}

@inproceedings{li2025lvr,
  title     = {Latent Visual Reasoning},
  author    = {Li, Bangzheng and Sun, Ximeng and Liu, Jiang and Wang, Ze and Wu, Jialian and Yu, Xiaodong and Chen, Hao and Barsoum, Emad and Chen, Muhao and Liu, Zicheng},
  booktitle = {International Conference on Learning Representations (ICLR)},
  year      = {2026}
}

@article{xiao2026realvr,
  title     = {Rethinking Latent Visual Reasoning: Grounding Latent Reasoning in Visual Evidence},
  author    = {Xiao, Xi and Zhao, Tianchen and Kim, Youngeun and Li, Zhuowei and Xu, Linghan and Wu, Jiaye and Zhang, Zheng and Xu, Xiang and Chen, Xuanbai and Tejani, Farhan and Zablocki, Jakub and Xu, Julia and Xing, Yifan},
  journal   = {arXiv preprint arXiv:2609.34563},
  year      = {2026}
}

@inproceedings{yang2025mindjourney,
  title     = {{MindJourney}: Test-Time Scaling with World Models for Spatial Reasoning},
  author    = {Yang, Yuncong and Liu, Jiageng and Zhang, Zheyuan and Zhou, Siyuan and Tan, Reuben and Yang, Jianwei and Du, Yilun and Gan, Chuang},
  booktitle = {Advances in Neural Information Processing Systems (NeurIPS)},
  year      = {2025}
}

@inproceedings{wang2025mindcube,
  title     = {{MindCube}: Spatial Mental Modeling from Limited Views},
  author    = {Wang, Qineng and Yin, Baiqiao and Zhang, Pingyue and Zhang, Jianshu and Wang, Kangrui and Wang, Zihan and Zhang, Jieyu and Chandrasegaran, Keshigeyan and Liu, Han and Krishna, Ranjay and Xie, Saining and Wu, Jiajun and Fei-Fei, Li and Li, Manling},
  booktitle = {International Conference on Learning Representations (ICLR)},
  year      = {2026}
}

@inproceedings{yang2024thinkinginspace,
  title     = {Thinking in Space: How Multimodal Large Language Models See, Remember, and Recall Spaces},
  author    = {Yang, Jihan and Yang, Shusheng and Gupta, Anjali W. and Han, Rilyn and Fei-Fei, Li and Xie, Saining},
  booktitle = {Proceedings of the IEEE/CVF Conference on Computer Vision and Pattern Recognition (CVPR)},
  pages     = {10632--10643},
  year      = {2025}
}

@inproceedings{kancheti2026cotdegrades,
  title     = {Chain-of-Thought Degrades Visual Spatial Reasoning Capabilities of Multimodal {LLMs}},
  author    = {Kancheti, Sai Srinivas and Kanade, Aditya Sanjiv and Balasubramanian, Vineeth N. and Ganu, Tanuja},
  booktitle = {Proceedings of the 64th Annual Meeting of the Association for Computational Linguistics (Volume 2: Short Papers)},
  pages     = {862--876},
  year      = {2026}
}

@inproceedings{chan2026yesbias,
  title     = {System-Mediated Attention Imbalances Make Vision-Language Models Say Yes},
  author    = {Chan, Tsan Tsai and Suresh, Varsha and Saha, Anisha and Hahn, Michael and Demberg, Vera},
  booktitle = {Findings of the Association for Computational Linguistics: ACL 2026},
  pages     = {38958--38979},
  year      = {2026}
}

@inproceedings{zhang2025spinbench,
  title     = {{SpinBench}: Perspective and Rotation as a Lens on Spatial Reasoning in {VLMs}},
  author    = {Zhang, Yuyou and Corcodel, Radu and Hori, Chiori and Cherian, Anoop and Zhao, Ding},
  booktitle = {International Conference on Learning Representations (ICLR)},
  year      = {2026}
}
}

\providecommand{\tbd}[1]{\textcolor{red}{[TBD: #1]}}
\clearpage
\appendix
\setcounter{page}{1}
\setcounter{section}{0}
\renewcommand{\thesection}{\Alph{section}}
\maketitlesupplementary

\section{Dataset Statistics}
\label{app:stats}

Geometric statistics of the 1{,}000-model VLM evaluation set and of the full BlueCAD-30K sample (30{,}237 models, including the 312 that failed rendering), computed from each STEP B-rep with the OpenCascade extractor behind the selection score (Table~\ref{tab:complexity_signals}). \emph{Simple} and \emph{complex} denote the prismatic and freeform subsets, as in the main text.

\paragraph{Face-count threshold.}
The geometric filter removes models with fewer than $k{=}5$ faces: below that the unfiltered pool is almost entirely single extrusions of convex polygons, whereas 5--15 faces already covers stepped blocks, flanged plates, and parts with through holes. Over the full sample the smallest model has 7 faces and 4.1\% have fewer than 15; in the VLM set the share is 4.4\% (0.8\% of the simple subset, 8.0\% of the complex one, mostly B-spline-dominated parts). Every model is a single solid.

\paragraph{Sample composition.}
Over the full dataset the score spans 11.9--88.6 (median 33.5, interquartile range 26.6--41.9), so most data sits between the two VLM classes. 76\% of models contain no freeform face; 65\% of all faces are planar and 22\% cylindrical, while B-spline faces make up 4.9\% of the sample but 51\% of the complex subset.

\paragraph{Class separation.}
The score ranges 12.2--18.2 for the simple subset and 59.1--87.3 for the complex one (medians 16.8 and 68.5) with no overlap, and the classes are separated on every weighted signal, not only on their sum: 99\% of simple models have no freeform face and all have a non-planar face ratio below 0.2, whereas 95\% of complex models exceed 0.6 and 99.6\% have a curved-edge ratio above 0.5. Face count does not separate them (interquartile ranges 19--32 and 26--63 overlap), nor do sphericity and aspect ratio, so the distinction is one of surface type and topology rather than of size or silhouette.

\begin{table}[htbp]
\centering
\caption{Selection-score signals, weights (sum to 100), and normalization to $[0,1]$. $r$: raw ratio for the named signal; $H$: Shannon entropy of the surface-type histogram, normalized by $\log(11)$; $c$: Wadell's sphericity, $c=\sqrt[3]{36\pi V^2}/A$, clipped to $[0,1]$. Weights are fixed manually: the freeform-face ratio as the clearest primitive-vs-organic separator, surface-type entropy after the information-theoretic measures of SHREC'21~\cite{arslan2021shrec} with the per-face classification of~\cite{jayaraman2021uvnet,lambourne2021brepnet}, and compactness as Wadell's sphericity~\cite{wadell1935volume} entered as $1-c$.}
\label{tab:complexity_signals}
\small
\begin{tabular}{lcl}
\toprule
Signal & Weight & Normalization \\
\midrule
Freeform face ratio   & 25 & $\min(1,\, 2r_{\text{freeform}})$ \\
Face count (log)      & 15 & $\min(1,\, \log_{10}(\text{faces})/3)$ \\
Non-planar face ratio & 15 & $r_{\text{curved face}}$ \\
Surface-type entropy  & 15 & $\min(1,\, 1.5\,H)$ \\
Inner boundary loops  & 12 & $\min(1,\, \text{loops}/12)$ \\
Curved edge ratio     & 10 & $r_{\text{curved edge}}$ \\
Shape compactness     & 8  & $1 - c$ \\
\bottomrule
\end{tabular}
\end{table}

\paragraph{Near-duplicates and validity.}
Grouping models by an exact topological signature (face, edge, and inner-loop counts, the 11-bin surface-type histogram, and bounding-box proportions to four decimals) finds no shared signature in the VLM set and only three pairs (0.02\%) in the full sample, so exact duplicates do not affect retrieval.

\begin{table}[htbp]
\centering
\footnotesize
\setlength{\tabcolsep}{5pt}
\caption{Per-model medians for the VLM subsets (Simple = prismatic, Complex = freeform) and the full sample ($n{=}30{,}237$); interquartile ranges are visible in Fig.~\ref{fig:stats_hist}. Bottom block: fraction of all faces per surface class.}
\label{tab:stats_summary}
\begin{tabular}{@{}lccc@{}}
\toprule
 & Simple & Complex & Full \\
\midrule
Faces                 & 23   & 42   & 37 \\
Edges                 & 120  & 209  & 190 \\
Inner loops           & 0    & 4    & 4 \\
Freeform face ratio   & 0.00 & 0.50 & 0.00 \\
Non-planar face ratio & 0.07 & 0.81 & 0.30 \\
Curved edge ratio     & 0.08 & 0.86 & 0.28 \\
Surface-type entropy  & 0.11 & 0.43 & 0.28 \\
Sphericity $c$        & 0.50 & 0.43 & 0.41 \\
Bbox aspect           & 5.0  & 2.2  & 4.2 \\
Selection score $s$   & 16.8 & 68.5 & 33.5 \\
\midrule
Plane faces           & 0.92 & 0.19 & 0.65 \\
Cylinder faces        & 0.07 & 0.19 & 0.22 \\
B-spline faces        & 0.00 & 0.51 & 0.05 \\
Other faces           & 0.01 & 0.11 & 0.08 \\
\bottomrule
\end{tabular}
\end{table}

\begin{figure*}[t]
\centering
\includegraphics[width=\textwidth]{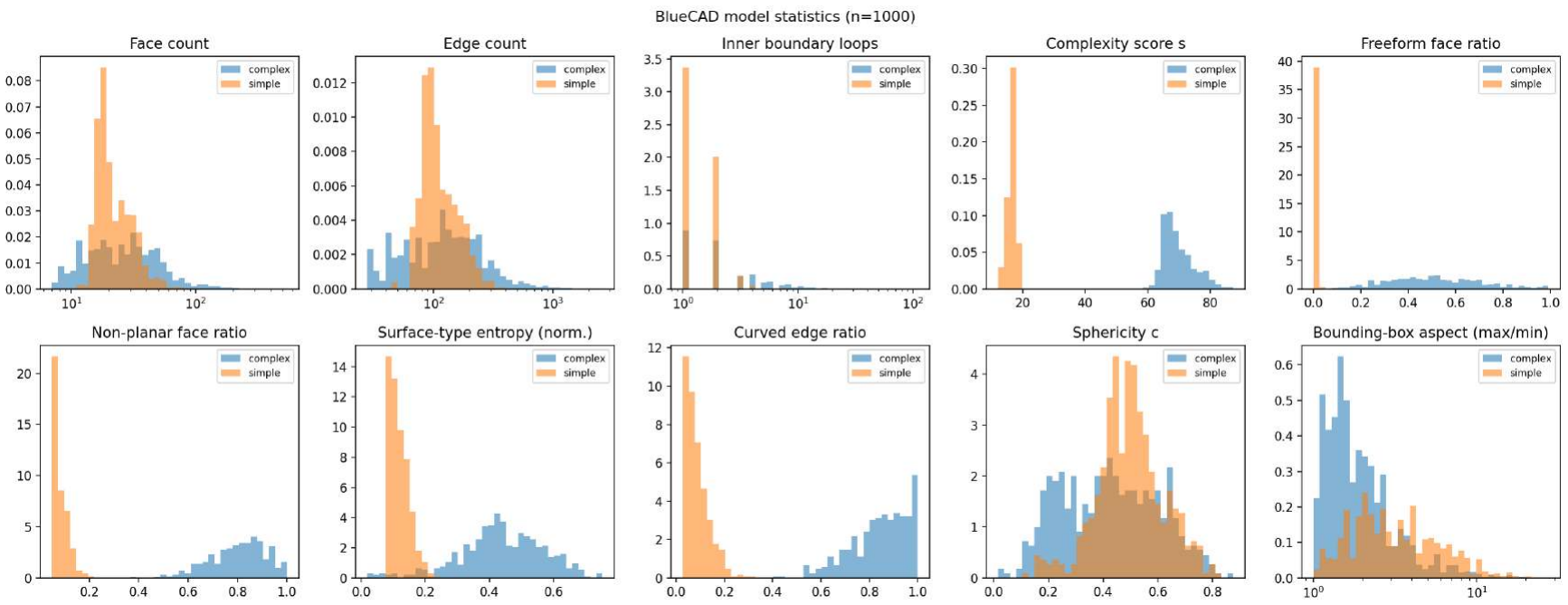}
\caption{Distributions of B-rep signals for the VLM evaluation subsets (simple in orange, complex in blue) against the full sample (grey). Face count, edge count, and aspect ratio are on a log axis.}
\label{fig:stats_hist}
\end{figure*}

\begin{figure}[htbp]
\centering
\includegraphics[width=\columnwidth]{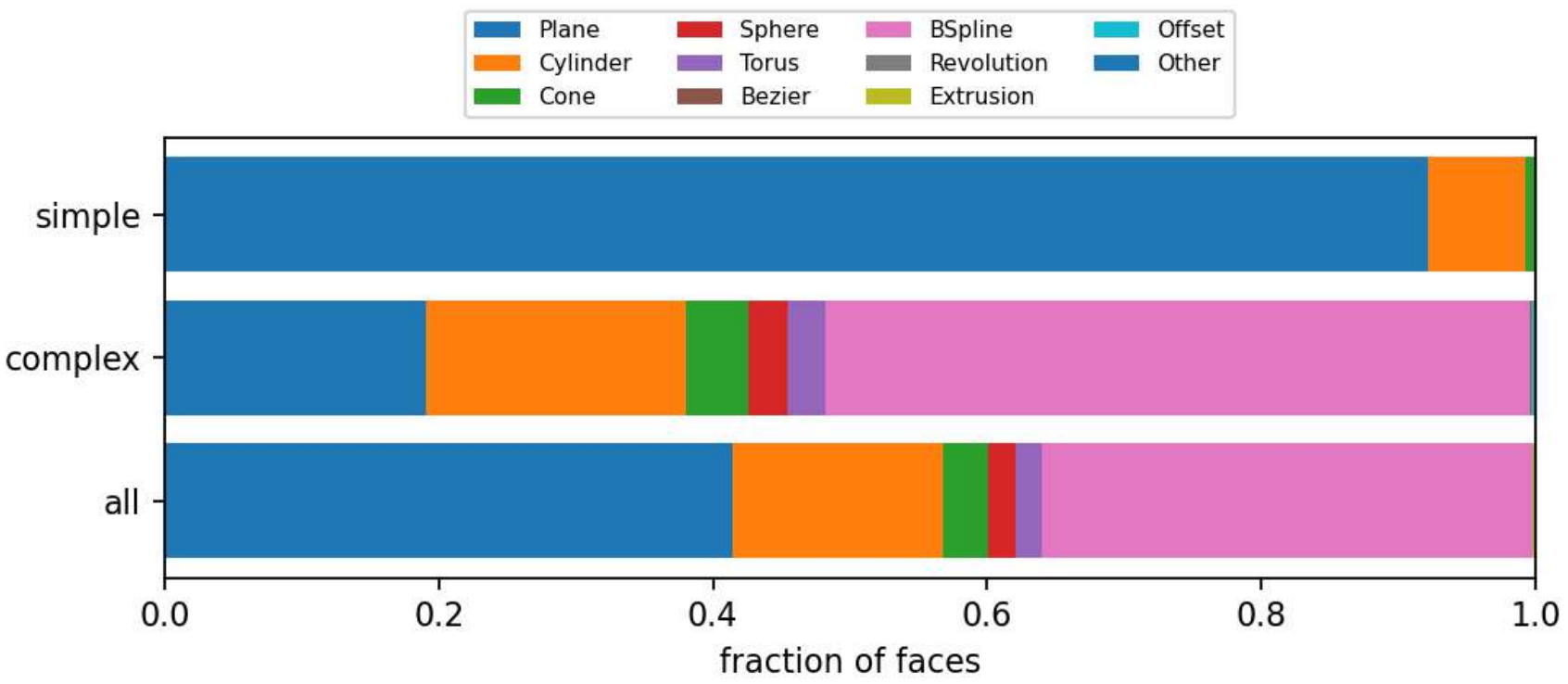}
\caption{Fraction of faces per surface type for the simple subset, the complex subset, and the full sample. Simple parts are almost entirely planar and cylindrical; complex parts are dominated by B-spline faces.}
\label{fig:surface_types}
\end{figure}

\begin{table}[htbp]
\centering
\caption{Canonical camera definitions. Projection directions point from the part toward the viewer (FreeCAD convention); up $=$ direction $\times$ $x$-axis. All vectors are in native STEP coordinates, with no per-model reorientation.}
\label{tab:cameras}
\begin{tabular}{lccc}
\toprule
View & Proj. direction & Image $x$-axis & Image up \\
\midrule
Front  & $(0,-1,0)$ & $(1,0,0)$  & $(0,0,1)$ \\
Right  & $(1,0,0)$  & $(0,1,0)$  & $(0,0,1)$ \\
Back   & $(0,1,0)$  & $(-1,0,0)$ & $(0,0,1)$ \\
Left   & $(-1,0,0)$ & $(0,-1,0)$ & $(0,0,1)$ \\
Top    & $(0,0,1)$  & $(1,0,0)$  & $(0,1,0)$ \\
Bottom & $(0,0,-1)$ & $(1,0,0)$  & $(0,-1,0)$ \\
\bottomrule
\end{tabular}
\end{table}

\paragraph{Edge treatment.}
Drawings use the \texttt{HLRBRep\_HLRToShape} edge sets: sharp (\texttt{V/HCompound}) and silhouette (\texttt{OutLineV/HCompound}) edges are drawn solid when visible and dashed when hidden, while tangent, seam, and isoparametric edges are suppressed, matching the TechDraw default.

\section{Frozen-Backbone Results}
\label{sec:appendix_frozen}

\paragraph{Checkpoints.} DINOv3 ViT-H+/16, SigLIP2 So400m/14 (384\,px), RADIO ViT-H/16, and TIPSv2 ViT-g/14 (448\,px). HF models use each checkpoint's default image processor (resize and center-crop); TIPSv2 images are resized directly to square. Renders are stored as RGBA with a transparent background and converted to RGB before encoding, so every encoder sees them on a black background, whereas drawings are on white. Global descriptors are the CLS token (DINOv3, TIPSv2), the pooled output (SigLIP2), or the summary token (RADIO), computed in fp16 autocast, mean-centered per modality over the gallery, and $\ell_2$-normalized. The \emph{Silhouette} rows use the model-free descriptor of Sec.~\ref{app:finetune}: the foreground mask of each view (drawings: outline closed and filled, so hidden and interior lines are discarded), cropped to its bounding box, resampled to $16{\times}16$, concatenated with log aspect ratio and fill fraction, and $\ell_2$-normalized.

\subsection{Full-Gallery Retrieval}
Tab.~\ref{tab:exp1_full} extends Tab.~\ref{tab:exp1} with all recall cutoffs, MRR, and median rank in both retrieval directions for the view-aligned (Al-mean) setting. All four encoders place the correct part first for most queries and within the top ten for 82--87\% of them. Each retrieves better from drawings to renders than the reverse at every cutoff, by the widest margin for RADIO, whereas the outline baseline is symmetric.

\begin{table}[htbp]
\centering
\caption{\textbf{Complete frozen-backbone retrieval results} (extends Tab.~\ref{tab:exp1}). View-aligned setting (Al-mean, viewpoints known), both directions, all metrics; \emph{Silhouette} is the non-learned outline baseline. Gallery: $N{=}10{,}000$-model sample; chance R@1 $= 0.01\%$, chance median rank $\approx 5{,}000$.}
\label{tab:exp1_full}
\resizebox{\linewidth}{!}{%
\begin{tabular}{llccccccc}
\toprule
\rowcolor{gray!20}
\textbf{Backbone} & \textbf{Direction} & \textbf{R@1} $\uparrow$ & \textbf{R@2} $\uparrow$ & \textbf{R@3} $\uparrow$ & \textbf{R@5} $\uparrow$ & \textbf{R@10} $\uparrow$ & \textbf{MRR} $\uparrow$ & \textbf{MedR} $\downarrow$ \\
\midrule
\multirow{2}{*}{Silhouette} & D$\rightarrow$R & 48.54 & 55.41 & 59.04 & 62.79 & 68.34 & 0.555 & 2 \\
                            & R$\rightarrow$D & 48.08 & 55.40 & 58.75 & 63.14 & 68.98 & 0.554 & 2 \\
\midrule
\multirow{2}{*}{DINOv3}     & D$\rightarrow$R & 65.14 & 73.56 & 77.67 & 81.79 & 86.87 & 0.728 & 1 \\
                            & R$\rightarrow$D & 57.57 & 66.34 & 70.32 & 75.28 & 81.44 & 0.659 & 1 \\
\midrule
\multirow{2}{*}{TIPSv2}     & D$\rightarrow$R & 64.65 & 73.34 & 77.08 & 81.63 & 86.42 & 0.724 & 1 \\
                            & R$\rightarrow$D & 58.67 & 66.60 & 71.11 & 75.55 & 81.18 & 0.665 & 1 \\
\midrule
\multirow{2}{*}{SigLIP2}    & D$\rightarrow$R & 64.15 & 71.99 & 75.50 & 79.71 & 84.50 & 0.714 & 1 \\
                            & R$\rightarrow$D & 57.15 & 65.17 & 69.29 & 73.61 & 79.27 & 0.649 & 1 \\
\midrule
\multirow{2}{*}{RADIO}      & D$\rightarrow$R & 59.51 & 67.86 & 72.37 & 77.01 & 82.36 & 0.675 & 1 \\
                            & R$\rightarrow$D & 49.58 & 57.71 & 62.16 & 66.65 & 73.19 & 0.578 & 2 \\
\bottomrule
\end{tabular}}
\end{table}

\subsection{Per-View Analysis}
Tab.~\ref{tab:exp2_full_all} gives single- and two-view results for all four encoders, and the pattern is the same for all of them. An opposing pair (front\,+\,back, left\,+\,right, top\,+\,bottom) improves on its stronger member by at most 2.5 R@1 points, whereas an orthogonal pair adds 11--22 points; \emph{top} is the strongest single view and \emph{bottom} the second, and top\,+\,left is the best pair. Information is thus spread across axes rather than across views, and the plan view carries the most of it.

\begin{table*}[htbp]
    \centering
    \caption{\textbf{Per-view analysis of the four encoders}. Drawing$\rightarrow$render, view-aligned mean pooling, gallery $N{=}10{,}000$. Single orthographic views (top block, sorted by R@1) and two-view fusion (bottom block). Best pair per backbone and metric in bold.}
    \label{tab:exp2_full_all}

    \begin{minipage}{0.48\textwidth}
    \centering \textbf{DINOv3}\\[2pt]
    \resizebox{\textwidth}{!}{%
    \begin{tabular}{llccccc}
    \toprule
    \rowcolor{gray!20}
    & \textbf{View} & \textbf{R@1}$\uparrow$ & \textbf{R@5}$\uparrow$ & \textbf{R@10}$\uparrow$ & \textbf{MRR}$\uparrow$ & \textbf{MedR}$\downarrow$ \\
    \midrule
    \multirow{6}{*}{\rotatebox{90}{Single}}
     & top    & 30.98 & 44.49 & 50.60 & 0.377 & 10 \\
     & bottom & 25.87 & 38.72 & 44.42 & 0.323 & 19 \\
     & left   & 16.53 & 28.20 & 33.95 & 0.225 & 50 \\
     & right  & 16.41 & 27.58 & 33.40 & 0.223 & 55 \\
     & front  & 14.88 & 25.98 & 31.38 & 0.206 & 62 \\
     & back   & 14.49 & 24.26 & 29.54 & 0.198 & 72 \\
    \cmidrule{1-7}
    \multirow{6}{*}{\rotatebox{90}{Pairs}}
     & top + left   & \textbf{53.01} & \textbf{72.35} & \textbf{78.75} & \textbf{0.621} & \textbf{1} \\
     & front + top  & 51.11 & 70.74 & 77.20 & 0.602 & \textbf{1} \\
     & top + bottom & 32.99 & 47.39 & 53.52 & 0.402 & 7 \\
     & front + left & 29.05 & 44.30 & 51.03 & 0.365 & 9 \\
     & left + right & 18.24 & 30.24 & 36.24 & 0.244 & 38 \\
     & front + back & 15.76 & 27.08 & 33.11 & 0.217 & 48 \\
    \bottomrule
    \end{tabular}}
    \end{minipage}
    \hfill
    \begin{minipage}{0.48\textwidth}
    \centering \textbf{TIPSv2}\\[2pt]
    \resizebox{\textwidth}{!}{%
    \begin{tabular}{llccccc}
    \toprule
    \rowcolor{gray!20}
    & \textbf{View} & \textbf{R@1}$\uparrow$ & \textbf{R@5}$\uparrow$ & \textbf{R@10}$\uparrow$ & \textbf{MRR}$\uparrow$ & \textbf{MedR}$\downarrow$ \\
    \midrule
    \multirow{6}{*}{\rotatebox{90}{Single}}
     & top    & 32.58 & 48.31 & 54.39 & 0.402 & 6 \\
     & bottom & 27.56 & 41.96 & 48.53 & 0.347 & 12 \\
     & left   & 15.71 & 27.10 & 32.36 & 0.216 & 58 \\
     & right  & 15.66 & 27.07 & 32.14 & 0.215 & 57 \\
     & front  & 13.32 & 24.14 & 30.18 & 0.191 & 71 \\
     & back   & 12.79 & 23.12 & 28.70 & 0.183 & 80 \\
    \cmidrule{1-7}
    \multirow{6}{*}{\rotatebox{90}{Pairs}}
     & top + left   & \textbf{54.22} & \textbf{74.20} & \textbf{80.66} & \textbf{0.634} & \textbf{1} \\
     & front + top  & 50.70 & 71.53 & 78.38 & 0.603 & \textbf{1} \\
     & top + bottom & 34.70 & 50.16 & 56.69 & 0.423 & 5 \\
     & front + left & 27.65 & 42.23 & 49.05 & 0.349 & 12 \\
     & left + right & 16.57 & 28.46 & 34.52 & 0.228 & 44 \\
     & front + back & 13.96 & 25.14 & 31.24 & 0.199 & 59 \\
    \bottomrule
    \end{tabular}}
    \end{minipage}

    \vspace{6pt}

    \begin{minipage}{0.48\textwidth}
    \centering \textbf{SigLIP2}\\[2pt]
    \resizebox{\textwidth}{!}{%
    \begin{tabular}{llccccc}
    \toprule
    \rowcolor{gray!20}
    & \textbf{View} & \textbf{R@1}$\uparrow$ & \textbf{R@5}$\uparrow$ & \textbf{R@10}$\uparrow$ & \textbf{MRR}$\uparrow$ & \textbf{MedR}$\downarrow$ \\
    \midrule
    \multirow{6}{*}{\rotatebox{90}{Single}}
     & top    & 30.82 & 46.06 & 52.32 & 0.383 & 8 \\
     & bottom & 23.68 & 37.42 & 43.49 & 0.305 & 20 \\
     & left   & 15.02 & 26.36 & 32.30 & 0.210 & 54 \\
     & right  & 14.85 & 26.62 & 32.44 & 0.209 & 53 \\
     & front  & 13.85 & 24.82 & 30.46 & 0.195 & 67 \\
     & back   & 12.53 & 23.49 & 29.55 & 0.183 & 76 \\
    \cmidrule{1-7}
    \multirow{6}{*}{\rotatebox{90}{Pairs}}
     & top + left   & \textbf{51.16} & \textbf{70.75} & \textbf{77.36} & \textbf{0.603} & \textbf{1} \\
     & front + top  & 49.83 & 69.10 & 76.10 & 0.589 & 2 \\
     & top + bottom & 33.32 & 49.40 & 56.22 & 0.412 & 6 \\
     & front + left & 28.22 & 43.72 & 50.34 & 0.358 & 10 \\
     & left + right & 16.19 & 28.70 & 35.38 & 0.227 & 40 \\
     & front + back & 14.35 & 26.60 & 32.59 & 0.206 & 54 \\
    \bottomrule
    \end{tabular}}
    \end{minipage}
    \hfill
    \begin{minipage}{0.48\textwidth}
    \centering \textbf{RADIO}\\[2pt]
    \resizebox{\textwidth}{!}{%
    \begin{tabular}{llccccc}
    \toprule
    \rowcolor{gray!20}
    & \textbf{View} & \textbf{R@1}$\uparrow$ & \textbf{R@5}$\uparrow$ & \textbf{R@10}$\uparrow$ & \textbf{MRR}$\uparrow$ & \textbf{MedR}$\downarrow$ \\
    \midrule
    \multirow{6}{*}{\rotatebox{90}{Single}}
     & top    & 29.95 & 44.06 & 50.42 & 0.370 & 10 \\
     & bottom & 23.34 & 36.07 & 42.19 & 0.297 & 25 \\
     & left   & 13.85 & 24.63 & 30.71 & 0.195 & 69 \\
     & right  & 13.44 & 24.41 & 30.26 & 0.192 & 69 \\
     & front  & 11.42 & 22.02 & 27.51 & 0.170 & 84 \\
     & back   & 10.96 & 21.42 & 26.53 & 0.164 & 96 \\
    \cmidrule{1-7}
    \multirow{6}{*}{\rotatebox{90}{Pairs}}
     & top + left   & \textbf{48.23} & \textbf{67.78} & \textbf{74.91} & \textbf{0.575} & \textbf{2} \\
     & front + top  & 45.62 & 66.28 & 73.30 & 0.552 & \textbf{2} \\
     & top + bottom & 29.96 & 44.85 & 51.52 & 0.374 & 9 \\
     & front + left & 25.23 & 40.70 & 46.57 & 0.326 & 15 \\
     & left + right & 14.28 & 25.77 & 31.85 & 0.203 & 54 \\
     & front + back & 11.93 & 23.02 & 28.71 & 0.178 & 71 \\
    \bottomrule
    \end{tabular}}
    \end{minipage}

\end{table*}

\subsection{Annotation Ablation}
Every gap between clean and dimensioned drawings narrows from R@1 to R@10 (DINOv3 7.0 to 5.3 points, TIPSv2 3.0 to 1.7, SigLIP2 2.6 to 0.7, RADIO 0.7 to $-0.6$), so annotations mostly demote the correct part from first place to a nearby rank rather than losing it. The exchange is one-sided only for DINOv3, which loses 1{,}132 first-place hits and gains 428; RADIO loses 862 and gains 795, a wash. Clean-drawing results for all cutoffs are in Tab.~\ref{tab:exp1_full}.

\section{Vision-Language Model Results}
\label{sec:appendix_vlm}

\subsection{Inference Settings}
\label{app:vlm_settings}
Tab.~\ref{tab:vlm_settings} lists the model identifiers and decoding settings. Closed models were queried through their vendors' APIs between 15 and 25 August 2026. No temperature, sampling, or reasoning parameter was set in any call, so each closed model ran with its provider's defaults, and the identifiers are the aliases passed to the API, which the vendor resolves to its snapshot current at that time. The zero-shot open models were served through an OpenAI-compatible local endpoint using the "vllm" library, and likewise ran with the server's default sampling, taken from each checkpoint's generation config. The fine-tuned model ran in-process with Hugging Face Transformers in bf16 with greedy decoding, so its comparison with zero-shot Qwen3-VL also changes the decoding. Every image is stored at $512{\times}512$\,px and sent to every model at that resolution, in the order given by each prompt. Responses without a parseable answer are excluded from that model's accuracy, and every paired comparison uses only the items both systems answered. The forced-choice runs contain one empty GPT-5.6~Luna response (Tab.~\ref{tab:task1_by_view_nodim}), and the CoT runs lose 41 GPT-5.6~Luna responses (empty), 11 Gemini responses (API errors), and 3 Claude responses (empty or cut off at the token limit); Tab.~\ref{tab:cot} reports the remaining $n$. The same rule removes 52 GPT-5.6~Luna and 7 Claude empty responses and 6 Gemini API errors from Task~2, and 4 Gemini API errors and 4 unparseable Mistral-S answers from Task~3; every other Task~2 and~3 run answered all 1{,}000 items. The reasoning setting matters for reading Tab.~\ref{tab:cot}: a model with internal reasoning enabled already reasons before its forced-choice answer, so for it the CoT control changes only whether that reasoning is written out and guided.

\begin{table}[htbp]
    \centering
    \caption{\textbf{VLM inference settings.} Identifier: API model alias (closed, queried 15--25 August 2026) or Hugging Face checkpoint (open). Temp.: no run set a temperature, so ``default'' is the provider's or the serving engine's default. Reasoning: ``off'' if thinking is disabled, ``default'' if left unset, n/a for models without a reasoning mode. Max tok.: output-token limit passed in the call, the same for forced-choice and CoT prompts; for GPT-5.6 Luna it also counts reasoning tokens.}
    \label{tab:vlm_settings}
    \footnotesize
    \setlength{\tabcolsep}{3pt}
    \resizebox{\columnwidth}{!}{%
    \begin{tabular}{lllll}
    \toprule
    \rowcolor{gray!20}
    \textbf{Model} & \textbf{Identifier} & \textbf{Temp.} & \textbf{Reasoning} & \textbf{Max tok.} \\
    \midrule
    Claude Sonnet 5   & \texttt{claude-sonnet-5}  & default & off     & 2000 \\
    Gemini 3.6 Flash  & \texttt{gemini-3.6-flash} & default & default & 2000 \\
    GPT-5.6 Luna      & \texttt{gpt-5.6-luna}     & default & default & 2000 \\
    \midrule
    Mistral-S 3.2 24B & \texttt{mistralai/Mistral-Small-3.2-24B-Instruct-2506} & default & n/a & 2000 (20 in T3) \\
    InternVL3.5 14B   & \texttt{OpenGVLab/InternVL3\_5-14B} & default & off & 2000 (100 in T3) \\
    Qwen3-VL 8B       & \texttt{Qwen/Qwen3-VL-8B-Instruct}  & default & n/a & 2000 \\
    Qwen3-VL 8B (FT)  & base + merged LoRA & greedy & n/a & 2000 \\
    \bottomrule
    \end{tabular}}
\end{table}

\subsection{Paired Significance Tests}
\label{app:paired_tests}
All \gls{vlm}s answer the same items, so accuracy differences are tested on paired outcomes rather than by overlapping per-model intervals. For two systems A and B, let $b$ be the items A answers correctly and B does not, and $c$ the reverse; the exact (binomial) McNemar test asks whether $b$ and $c$ differ, using only the $b+c$ discordant items, and $\Delta\text{Acc}=(b-c)/n$. Each comparison uses the $n$ items both systems answered (Sec.~\ref{app:vlm_settings}), so $\Delta$Acc can differ from the difference of Tab.~\ref{tab:main_results} entries by 0.001--0.002. $p$-values are Holm-corrected within each block of Tab.~\ref{tab:paired_tests}; the leading-model blocks list the four highest-accuracy zero-shot models of each condition. Over all 15 zero-shot pairs per condition, Mistral-S is below every other model ($p_{\text{Holm}}<0.001$) except Claude with dimensions ($p_{\text{Holm}}=0.089$), and with dimensions Claude is below Gemini, GPT-5.6, and Qwen3-VL ($p_{\text{Holm}}\le0.010$); no other pair differs.

\begin{table}[htbp]
    \centering
    \caption{\textbf{Paired exact McNemar tests} on the items both systems answered. $\Delta$Acc: accuracy of the first system minus the second on those items; $b{+}c$: discordant items; $p_{\text{Holm}}$: Holm-corrected within each block.}
    \label{tab:paired_tests}
    \footnotesize
    \setlength{\tabcolsep}{4pt}
    \begin{tabular}{lrrr}
    \toprule
    \rowcolor{gray!20}
    \textbf{Comparison} & $\Delta$\textbf{Acc} & $b{+}c$ & $p_{\text{Holm}}$ \\
    \midrule
    \multicolumn{4}{l}{\textit{Task 1, leading models (undimensioned)}} \\
    GPT-5.6 vs.\ Qwen3-VL   & $+0.000$ & 342 & 1.000 \\
    GPT-5.6 vs.\ Gemini     & $+0.017$ & 431 & 1.000 \\
    GPT-5.6 vs.\ Claude     & $+0.036$ & 344 & 0.295 \\
    Qwen3-VL vs.\ Gemini    & $+0.016$ & 418 & 1.000 \\
    Qwen3-VL vs.\ Claude    & $+0.035$ & 295 & 0.285 \\
    Gemini vs.\ Claude      & $+0.019$ & 497 & 1.000 \\
    \midrule
    \multicolumn{4}{l}{\textit{Task 1, leading models (dimensioned)}} \\
    Gemini vs.\ Qwen3-VL    & $+0.010$ & 378 & 1.000 \\
    Gemini vs.\ GPT-5.6     & $+0.016$ & 470 & 1.000 \\
    Gemini vs.\ InternVL3.5 & $+0.033$ & 307 & 0.406 \\
    Qwen3-VL vs.\ GPT-5.6   & $+0.006$ & 400 & 1.000 \\
    Qwen3-VL vs.\ InternVL3.5 & $+0.023$ & 277 & 0.931 \\
    GPT-5.6 vs.\ InternVL3.5 & $+0.017$ & 443 & 1.000 \\
    \midrule
    \multicolumn{4}{l}{\textit{Task 1, dimensioned vs.\ undimensioned}} \\
    Claude Sonnet 5   & $-0.045$ & 227 & 0.020 \\
    Gemini 3.6 Flash  & $+0.016$ & 148 & 1.000 \\
    GPT-5.6 Luna      & $-0.016$ & 254 & 1.000 \\
    Mistral-S 3.2 24B & $-0.010$ & 124 & 1.000 \\
    InternVL3.5 14B   & $+0.013$ & 217 & 1.000 \\
    Qwen3-VL 8B       & $-0.010$ & 148 & 1.000 \\
    \midrule
    \multicolumn{4}{l}{\textit{Task 1, CoT vs.\ forced choice (undimensioned)}} \\
    Claude Sonnet 5   & $+0.038$ & 302 & 0.165 \\
    Gemini 3.6 Flash  & $+0.014$ & 134 & 1.000 \\
    GPT-5.6 Luna      & $+0.007$ & 277 & 1.000 \\
    Mistral-S 3.2 24B & $+0.011$ & 673 & 1.000 \\
    InternVL3.5 14B   & $-0.007$ & 195 & 1.000 \\
    Qwen3-VL 8B       & $-0.049$ & 221 & 0.007 \\
    \midrule
    \multicolumn{4}{l}{\textit{Qwen3-VL 8B, fine-tuned vs.\ zero-shot (Task 1)}} \\
    Task 1            & $+0.016$ & 300 & 0.387 \\
    Task 1 + DIM      & $+0.040$ & 260 & 0.031 \\
    \midrule
    \multicolumn{4}{l}{\textit{Qwen3-VL 8B, fine-tuned vs.\ zero-shot (Tasks 2 and 3)}} \\
    Task 2            & $-0.076$ & 314 & $<0.001$ \\
    Task 3            & $+0.041$ & 143 & $<0.001$ \\
    \bottomrule
    \end{tabular}
\end{table}

\subsection{Task 1 by Surface-Type Class and Tier}
Tab.~\ref{tab:task1_results} splits Task~1 by surface-type class and pair tier. Every model, the fine-tuned one included, scores 2--13 points higher on freeform than on prismatic parts. The tiers expose the response biases: Gemini, InternVL3.5, and Mistral-S reject negatives at 0.82--0.99 but accept positives at only 0.09--0.42, while Claude and GPT-5.6 show the reverse, most strongly with dimensions (Claude: 0.85 on positives, 0.33 on negatives). Same-class negatives stay within 0.07 of any-class negatives for all models, including the fine-tuned one despite its silhouette-mined training negatives (Sec.~\ref{app:finetune}); since they share only the coarse class, not overall shape, this does not test look-alike parts.

\begin{table}[htbp]
    \centering
    \caption{Task 1 VLM results by surface-type class and pair tier, without (left) and with (right) dimension annotations. Bold: best zero-shot model per column; the fine-tuned (FT) row is excluded.}
    \label{tab:task1_results}
    \footnotesize
    \setlength{\tabcolsep}{3pt}
    \resizebox{\columnwidth}{!}{%
    \begin{tabular}{llcccc|cccc}
    \toprule
    \rowcolor{gray!20}
     & & \multicolumn{4}{c|}{\textbf{Without DIM}} & \multicolumn{4}{c}{\textbf{With DIM}} \\
    \rowcolor{gray!20}
     & \textbf{Model} & Acc. & F1 & Rec. & Prec. & Acc. & F1 & Rec. & Prec. \\
    \midrule
    \multirow{7}{*}{\rotatebox{90}{Complex}}
     & Claude Sonnet 5   & 0.665 & 0.665 & 0.666 & 0.668 & 0.611 & 0.597 & 0.615 & 0.637 \\
     & Gemini 3.6 Flash  & \textbf{0.717} & 0.699 & \textbf{0.713} & \textbf{0.777} & \textbf{0.735} & \textbf{0.724} & \textbf{0.732} & \textbf{0.773} \\
     & GPT-5.6 Luna      & 0.711 & \textbf{0.710} & 0.710 & 0.711 & 0.705 & 0.704 & 0.707 & 0.711 \\
     & Mistral-S 3.2 24B & 0.545 & 0.428 & 0.537 & 0.675 & 0.563 & 0.488 & 0.556 & 0.628 \\
     & InternVL3.5 14B   & 0.659 & 0.638 & 0.655 & 0.698 & 0.663 & 0.639 & 0.659 & 0.711 \\
     & Qwen3-VL 8B       & 0.679 & 0.668 & 0.676 & 0.702 & 0.689 & 0.677 & 0.686 & 0.715 \\
     & Qwen3-VL 8B (FT)  & 0.712 & 0.694 & 0.707 & 0.763 & 0.716 & 0.706 & 0.713 & 0.742 \\
    \midrule
    \multirow{7}{*}{\rotatebox{90}{Simple}}
     & Claude Sonnet 5   & 0.604 & 0.595 & 0.601 & 0.610 & 0.568 & 0.516 & 0.562 & 0.604 \\
     & Gemini 3.6 Flash  & 0.590 & 0.534 & 0.597 & \textbf{0.697} & 0.604 & 0.561 & 0.610 & \textbf{0.689} \\
     & GPT-5.6 Luna      & 0.630 & 0.626 & 0.628 & 0.633 & 0.602 & 0.589 & 0.599 & 0.611 \\
     & Mistral-S 3.2 24B & 0.522 & 0.406 & 0.531 & 0.670 & 0.484 & 0.413 & 0.491 & 0.482 \\
     & InternVL3.5 14B   & 0.588 & 0.570 & 0.592 & 0.613 & 0.610 & 0.597 & 0.614 & 0.634 \\
     & Qwen3-VL 8B       & \textbf{0.660} & \textbf{0.658} & \textbf{0.659} & 0.661 & \textbf{0.630} & \textbf{0.629} & \textbf{0.631} & 0.633 \\
     & Qwen3-VL 8B (FT)  & 0.658 & 0.635 & 0.663 & 0.724 & 0.682 & 0.670 & 0.686 & 0.723 \\
    \midrule
     & & \textbf{Same-cl.\ n.} & \textbf{Any-cl.\ n.} & \textbf{Pos.} & & \textbf{Same-cl.\ n.} & \textbf{Any-cl.\ n.} & \textbf{Pos.} & \\
    \cmidrule(lr){3-5}\cmidrule(lr){7-9}
    \multirow{7}{*}{\rotatebox{90}{Tier acc.}}
     & Claude Sonnet 5   & 0.562 & 0.520 & \textbf{0.728} & & 0.329 & 0.332 & \textbf{0.848} & \\
     & Gemini 3.6 Flash  & 0.952 & 0.952 & 0.356 & & \textbf{0.924} & \textbf{0.932} & 0.412 & \\
     & GPT-5.6 Luna      & 0.655 & 0.620 & 0.703 & & 0.554 & 0.516 & 0.772 & \\
     & Mistral-S 3.2 24B & \textbf{0.976} & \textbf{0.988} & 0.086 & & 0.876 & 0.904 & 0.158 & \\
     & InternVL3.5 14B   & 0.815 & 0.880 & 0.400 & & 0.863 & 0.852 & 0.416 & \\
     & Qwen3-VL 8B       & 0.695 & 0.764 & 0.610 & & 0.767 & 0.792 & 0.540 & \\
     & Qwen3-VL 8B (FT)  & 0.924 & 0.936 & 0.440 & & 0.856 & 0.920 & 0.510 & \\
    \bottomrule
    \end{tabular}}
\end{table}

\subsection{Task 1: Render Viewpoint}
\label{sec:by_view}
Tabs.~\ref{tab:task1_by_view} and~\ref{tab:task1_by_view_nodim} split Task~1 by render viewpoint. Positive-pair accuracy is highest on the side renders and, for every model but Gemini (lowest on \textit{back}), lowest on \textit{top} and \textit{bottom}, where negative-pair accuracy is often higher than elsewhere (Claude: 0.72 on \textit{top} against 0.48--0.49 on the side views without dimensions), so part of the drop on positives is a shift toward \emph{No} on these views.

\begin{table}[htbp]
    \centering
    \caption{Task 1 accuracy by render viewpoint, with DIM, under the forced-choice Task~1 prompt of Sec.~\ref{sec:vision_language_models} (weighting by $n$ recovers the positive and negative accuracies of Tab.~\ref{tab:bias_dim}). Bold: best viewpoint per model; the fine-tuned (FT) row is not marked.}
    \label{tab:task1_by_view}
    \small
    \setlength{\tabcolsep}{4pt}
    \begin{tabular}{lcccccc}
    \toprule \rowcolor{gray!20}
    \textbf{Model} & \textbf{front} & \textbf{right} & \textbf{back} & \textbf{left} & \textbf{top} & \textbf{bottom} \\
    \midrule
    \multicolumn{7}{l}{\textit{Positive pairs} ($n$ = 86, 81, 92, 79, 78, 84)} \\
    Claude Sonnet 5   & 0.92 & \textbf{1.00} & 0.89 & \textbf{1.00} & 0.53 & 0.74 \\
    Gemini 3.6 Flash  & 0.35 & \textbf{0.56} & 0.27 & 0.54 & 0.42 & 0.36 \\
    GPT-5.6 Luna      & 0.76 & \textbf{0.95} & 0.74 & 0.89 & 0.63 & 0.68 \\
    Mistral-S 3.2 24B & 0.16 & 0.21 & 0.20 & \textbf{0.28} & 0.06 & 0.04 \\
    InternVL3.5 14B   & 0.43 & \textbf{0.68} & 0.42 & 0.66 & 0.14 & 0.17 \\
    Qwen3-VL 8B       & 0.55 & 0.77 & 0.60 & \textbf{0.89} & 0.19 & 0.25 \\
    Qwen3-VL 8B (FT)  & 0.49 & 0.78 & 0.51 & 0.81 & 0.22 & 0.26 \\
    \midrule
    \multicolumn{7}{l}{\textit{Negative pairs} ($n$ = 73, 89, 85, 84, 89, 79)} \\
    Claude Sonnet 5   & 0.30 & 0.29 & 0.31 & 0.24 & \textbf{0.44} & 0.41 \\
    Gemini 3.6 Flash  & \textbf{0.95} & 0.93 & 0.93 & 0.92 & 0.94 & 0.90 \\
    GPT-5.6 Luna      & 0.52 & 0.56 & \textbf{0.61} & 0.40 & 0.54 & 0.57 \\
    Mistral-S 3.2 24B & 0.84 & 0.88 & 0.86 & 0.87 & 0.92 & \textbf{0.97} \\
    InternVL3.5 14B   & 0.82 & 0.83 & 0.84 & 0.83 & \textbf{0.91} & \textbf{0.91} \\
    Qwen3-VL 8B       & 0.82 & 0.72 & 0.74 & 0.67 & \textbf{0.89} & 0.85 \\
    Qwen3-VL 8B (FT)  & 0.86 & 0.94 & 0.93 & 0.88 & 0.84 & 0.86 \\
    \bottomrule
    \end{tabular}
\end{table}

\begin{table}[htbp]
    \centering
    \caption{Task 1 accuracy by render viewpoint, without DIM, under the forced-choice Task~1 prompt of Sec.~\ref{sec:vision_language_models} (weighting by $n$ recovers the positive and negative accuracies of Tab.~\ref{tab:bias_dim}). Bold: best viewpoint per model; the fine-tuned (FT) row is not marked.}
    \label{tab:task1_by_view_nodim}
    \small
    \setlength{\tabcolsep}{4pt}
    \begin{tabular}{lcccccc}
    \toprule \rowcolor{gray!20}
    \textbf{Model} & \textbf{front} & \textbf{right} & \textbf{back} & \textbf{left} & \textbf{top} & \textbf{bottom} \\
    \midrule
    \multicolumn{7}{l}{\textit{Positive pairs} ($n$ = 86, 81, 92, 79, 78, 84/83$^\dagger$)} \\
    Claude Sonnet 5   & 0.77 & \textbf{0.98} & 0.68 & 0.97 & 0.46 & 0.51 \\
    Gemini 3.6 Flash  & 0.31 & \textbf{0.47} & 0.18 & \textbf{0.47} & 0.42 & 0.31 \\
    GPT-5.6 Luna      & 0.74 & \textbf{0.88} & 0.67 & 0.85 & 0.51 & 0.57 \\
    Mistral-S 3.2 24B & \textbf{0.12} & \textbf{0.12} & 0.10 & 0.11 & 0.06 & 0.00 \\
    InternVL3.5 14B   & 0.41 & 0.65 & 0.39 & \textbf{0.71} & 0.12 & 0.13 \\
    Qwen3-VL 8B       & 0.66 & 0.89 & 0.67 & \textbf{0.91} & 0.21 & 0.31 \\
    Qwen3-VL 8B (FT)  & 0.37 & 0.70 & 0.47 & 0.77 & 0.14 & 0.19 \\
    \midrule
    \multicolumn{7}{l}{\textit{Negative pairs} ($n$ = 73, 89, 85, 84, 89, 79)} \\
    Claude Sonnet 5   & 0.48 & 0.49 & 0.49 & 0.48 & \textbf{0.72} & 0.57 \\
    Gemini 3.6 Flash  & 0.97 & 0.96 & 0.96 & \textbf{0.98} & 0.93 & 0.91 \\
    GPT-5.6 Luna      & 0.58 & \textbf{0.70} & 0.69 & 0.57 & 0.64 & 0.63 \\
    Mistral-S 3.2 24B & 0.99 & 0.99 & 0.95 & 0.98 & \textbf{1.00} & 0.99 \\
    InternVL3.5 14B   & 0.89 & 0.85 & 0.82 & 0.74 & \textbf{0.90} & 0.89 \\
    Qwen3-VL 8B       & 0.70 & 0.69 & 0.73 & 0.62 & \textbf{0.83} & 0.81 \\
    Qwen3-VL 8B (FT)  & 0.95 & 0.94 & 0.92 & 0.90 & 0.90 & 0.97 \\
    \bottomrule
    \end{tabular}
    \vspace{2pt}
    {\footnotesize $^\dagger$GPT-5.6 Luna: $n{=}83$ for \textit{bottom} positives due to one empty response.}
\end{table}

\subsection{Task 2 by Surface-Type Class}
Tab.~\ref{tab:task2_results} splits Task~2 by surface-type class. Claude, Gemini, GPT-5.6, and Mistral-S do better on freeform parts, InternVL3.5 and Qwen3-VL marginally better on prismatic ones. Against chance (one-sided exact binomial test, Holm-corrected over the six zero-shot models), Gemini, GPT-5.6, Claude, and Qwen3-VL score above 0.25 ($p_{\text{Holm}}<10^{-5}$) and InternVL3.5 and Mistral-S do not ($p_{\text{Holm}}\ge0.088$); on the items both answered, Gemini leads GPT-5.6 by 6.3 points (paired McNemar $p<0.001$). The fine-tuned Qwen3-VL drops to chance on both classes (overall 0.241, $p=0.53$ against 0.25), 7.6 points below its zero-shot base ($p_{\text{Holm}}<0.001$; Tab.~\ref{tab:paired_tests}): it treats the drawing as the first labelled candidate, answering \emph{E} on 12.9\% of items and the letter one past the ground truth on 54.1\%.

\begin{figure}[htbp]
    \centering
    \includegraphics[width=\linewidth]{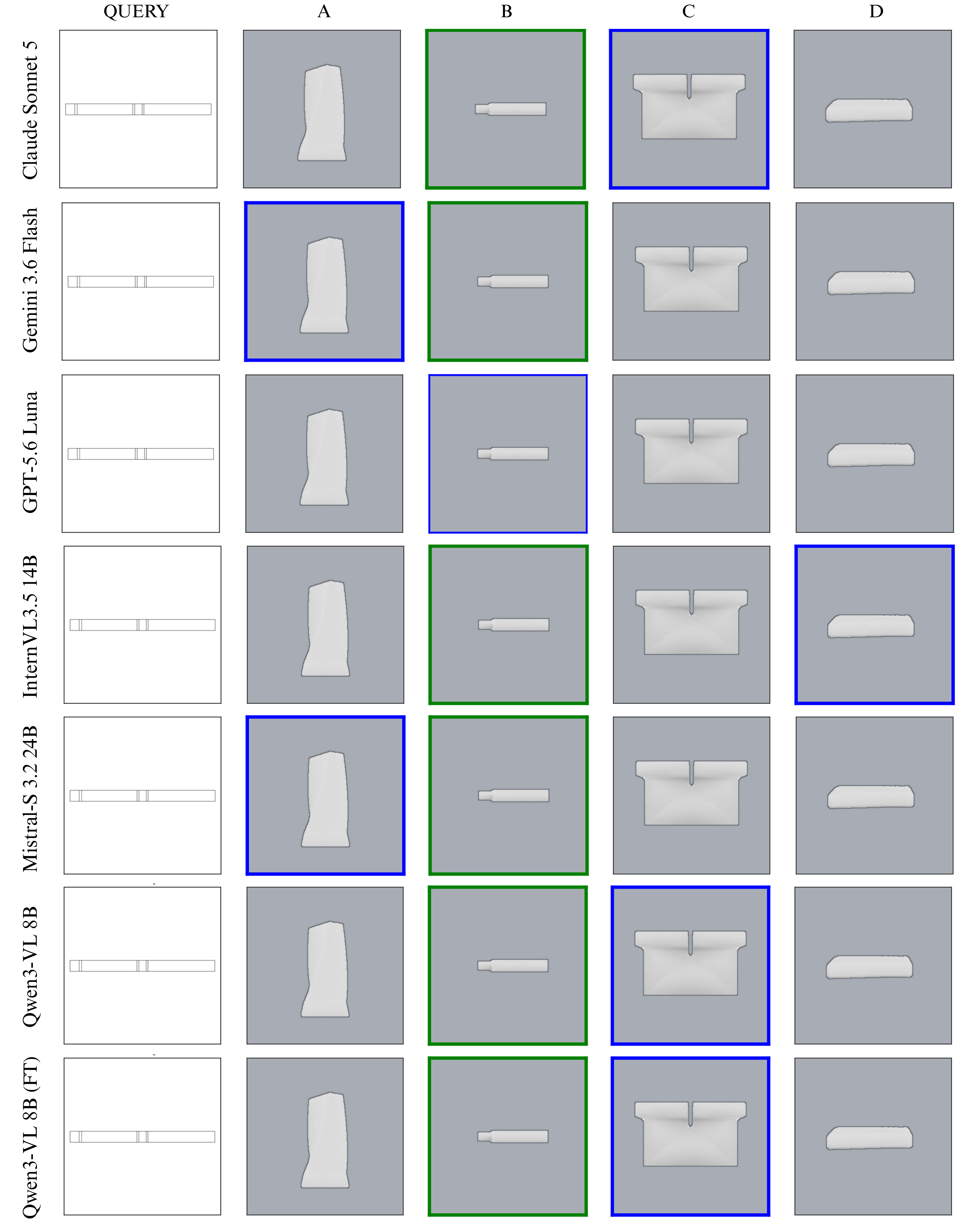}
    \caption{Examples of 2D drawing retrieval tasks with four candidate CAD renders. Green border: ground truth; blue border: model selection.}
    \label{fig:task2_samples_appex}
\end{figure}

\begin{table}[htbp]
    \centering
    \caption{Task 2 VLM results by model (four-way drawing-to-render retrieval). Bold: best zero-shot model per column; the fine-tuned (FT) row is excluded}
    \label{tab:task2_results}
    \small
    \setlength{\tabcolsep}{4pt}
    \begin{tabular}{llcccc}
    \toprule
    \rowcolor{gray!20}
     & \textbf{Model} & \textbf{Acc.} & \textbf{F1} & \textbf{Rec.} & \textbf{Prec.} \\
    \midrule
    \multirow{7}{*}{\rotatebox{90}{Overall}}
     & Claude Sonnet 5   & 0.374 & 0.372 & 0.379 & 0.425 \\
     & Gemini 3.6 Flash  & \textbf{0.690} & \textbf{0.689} & \textbf{0.689} & \textbf{0.691} \\
     & GPT-5.6 Luna      & 0.637 & 0.636 & 0.635 & 0.643 \\
     & Mistral-S 3.2 24B & 0.274 & 0.235 & 0.267 & 0.282 \\
     & InternVL3.5 14B   & 0.266 & 0.274 & 0.269 & 0.310 \\
     & Qwen3-VL 8B       & 0.317 & 0.305 & 0.321 & 0.322 \\
     & Qwen3-VL 8B (FT)  & 0.241 & 0.246 & 0.239 & 0.257 \\
    \midrule
    \multirow{7}{*}{\rotatebox{90}{Complex}}
     & Claude Sonnet 5   & 0.415 & 0.415 & 0.413 & 0.459 \\
     & Gemini 3.6 Flash  & \textbf{0.817} & \textbf{0.817} & \textbf{0.816} & \textbf{0.817} \\
     & GPT-5.6 Luna      & 0.740 & 0.740 & 0.739 & 0.744 \\
     & Mistral-S 3.2 24B & 0.284 & 0.250 & 0.285 & 0.288 \\
     & InternVL3.5 14B   & 0.264 & 0.276 & 0.264 & 0.309 \\
     & Qwen3-VL 8B       & 0.316 & 0.307 & 0.317 & 0.322 \\
     & Qwen3-VL 8B (FT)  & 0.236 & 0.241 & 0.233 & 0.256 \\
    \midrule
    \multirow{7}{*}{\rotatebox{90}{Simple}}
     & Claude Sonnet 5   & 0.332 & 0.326 & 0.343 & 0.389 \\
     & Gemini 3.6 Flash  & \textbf{0.564} & \textbf{0.563} & \textbf{0.563} & \textbf{0.566} \\
     & GPT-5.6 Luna      & 0.528 & 0.521 & 0.522 & 0.531 \\
     & Mistral-S 3.2 24B & 0.264 & 0.221 & 0.250 & 0.293 \\
     & InternVL3.5 14B   & 0.268 & 0.274 & 0.274 & 0.320 \\
     & Qwen3-VL 8B       & 0.318 & 0.303 & 0.325 & 0.325 \\
     & Qwen3-VL 8B (FT)  & 0.246 & 0.250 & 0.244 & 0.260 \\
    \bottomrule
    \end{tabular}
\end{table}

\subsection{Task 3: Complexity Probe}
\begin{figure*}[htbp]
    \centering
    \begin{minipage}[t]{0.49\linewidth}
        \centering
        \includegraphics[width=\linewidth]{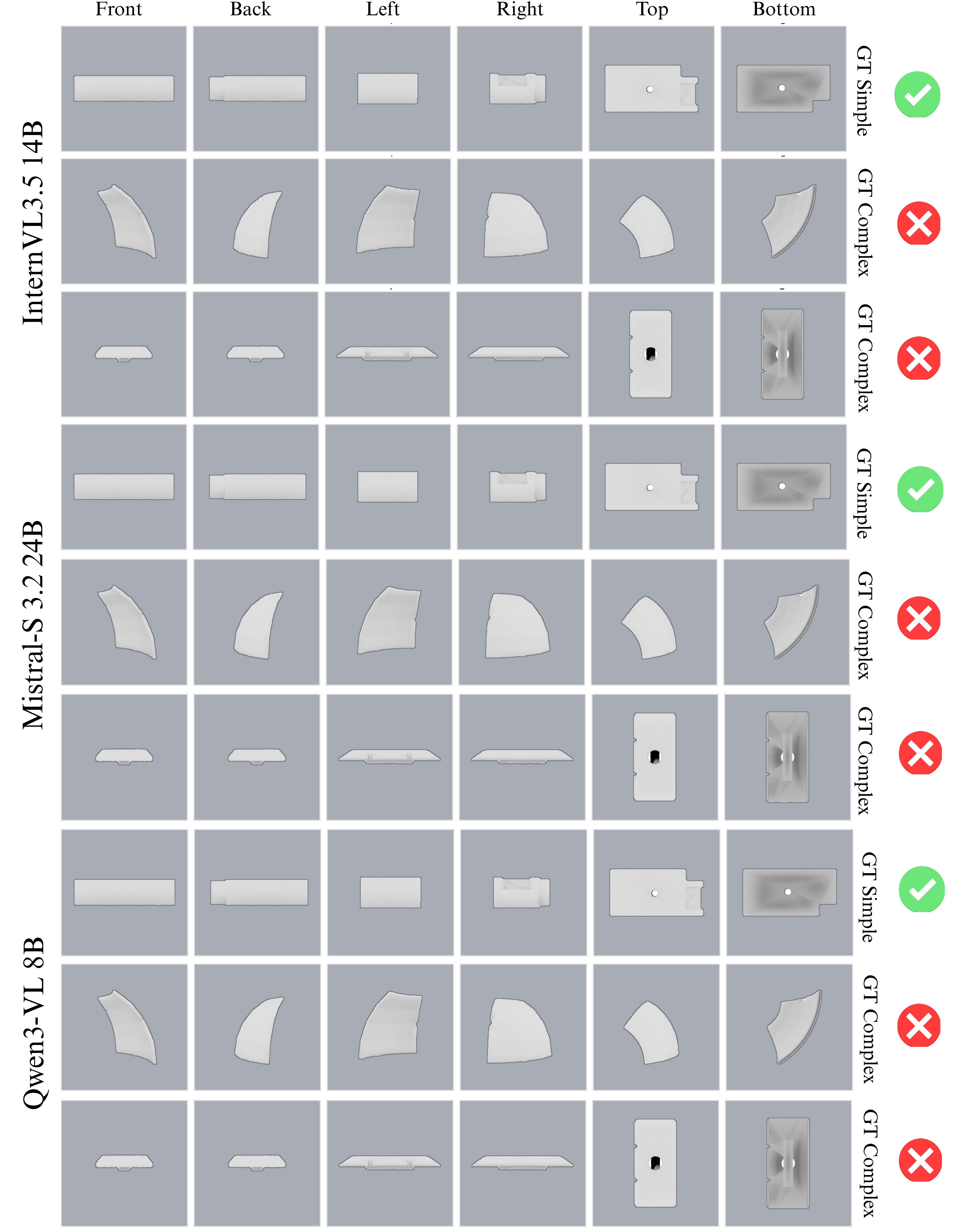}
        \subcaption{Open-source models.}
        \label{fig:task3_open_source}
    \end{minipage}
    \hfill
    \begin{minipage}[t]{0.49\linewidth}
        \centering
        \includegraphics[width=\linewidth]{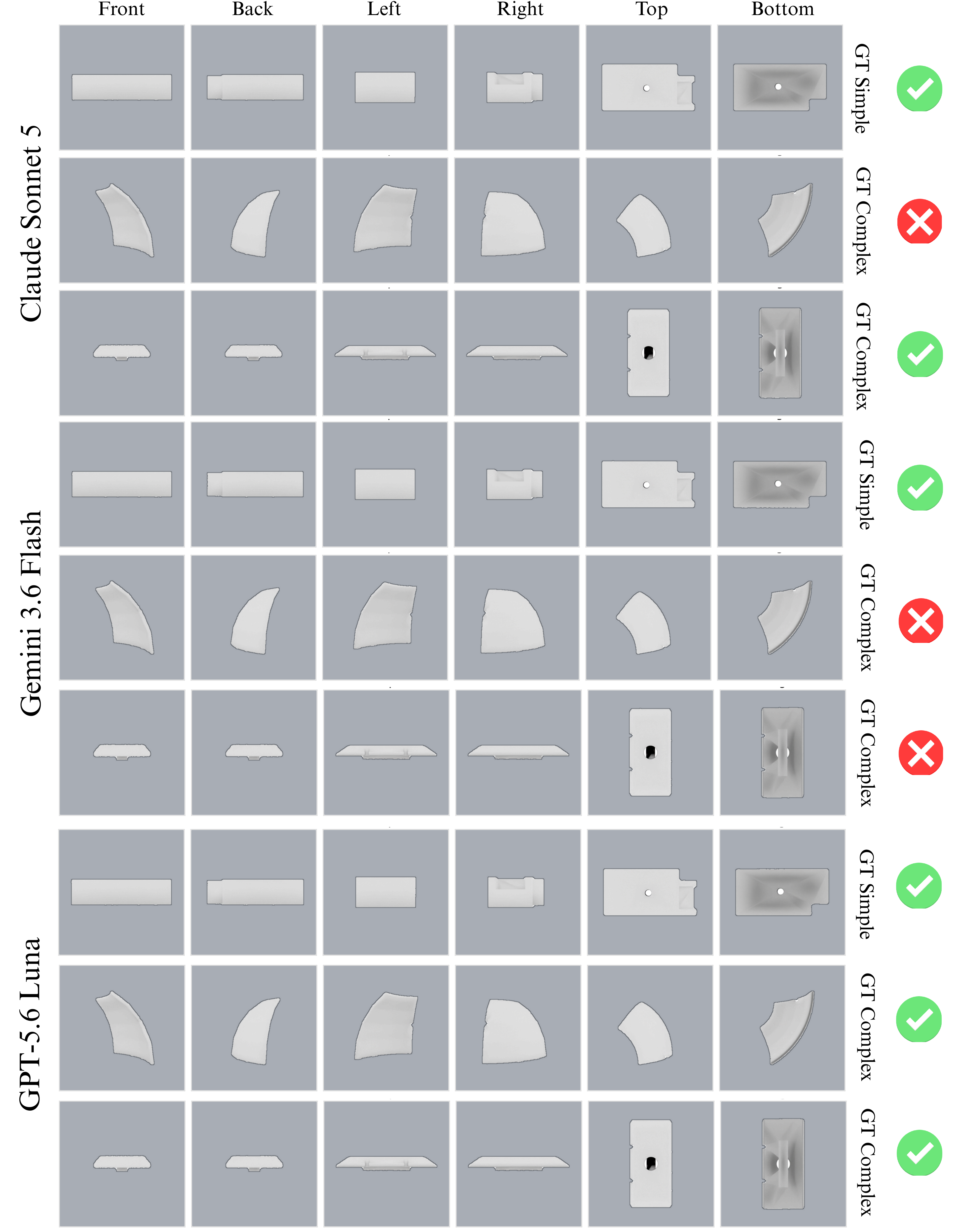}
        \subcaption{Closed-source models.}
        \label{fig:task3_closed_source}
    \end{minipage}
    \caption{Task 3 predictions across models (green = correct, red = incorrect).}
    \label{fig:sample_task3_models_appex}
\end{figure*}

On a balanced set, every model but GPT-5.6 Luna (46.4\%) answers \emph{Simple} on 87.6--98.4\% of items (Claude 93.4\%, Gemini 87.6\%, Mistral-S 93.4\%, InternVL3.5 92.2\%, Qwen3-VL 98.4\%). Fine-tuning on Task~1 moves Qwen3-VL from 98.4\% to 84.9\% \emph{Simple} answers (accuracy 0.512 to 0.553). Fig.~\ref{fig:sample_task3_models_appex} shows qualitative samples.

\subsection{Task 1 Failure Analysis}
\begin{figure*}[p]
    \centering
    \includegraphics[width=0.5\linewidth]{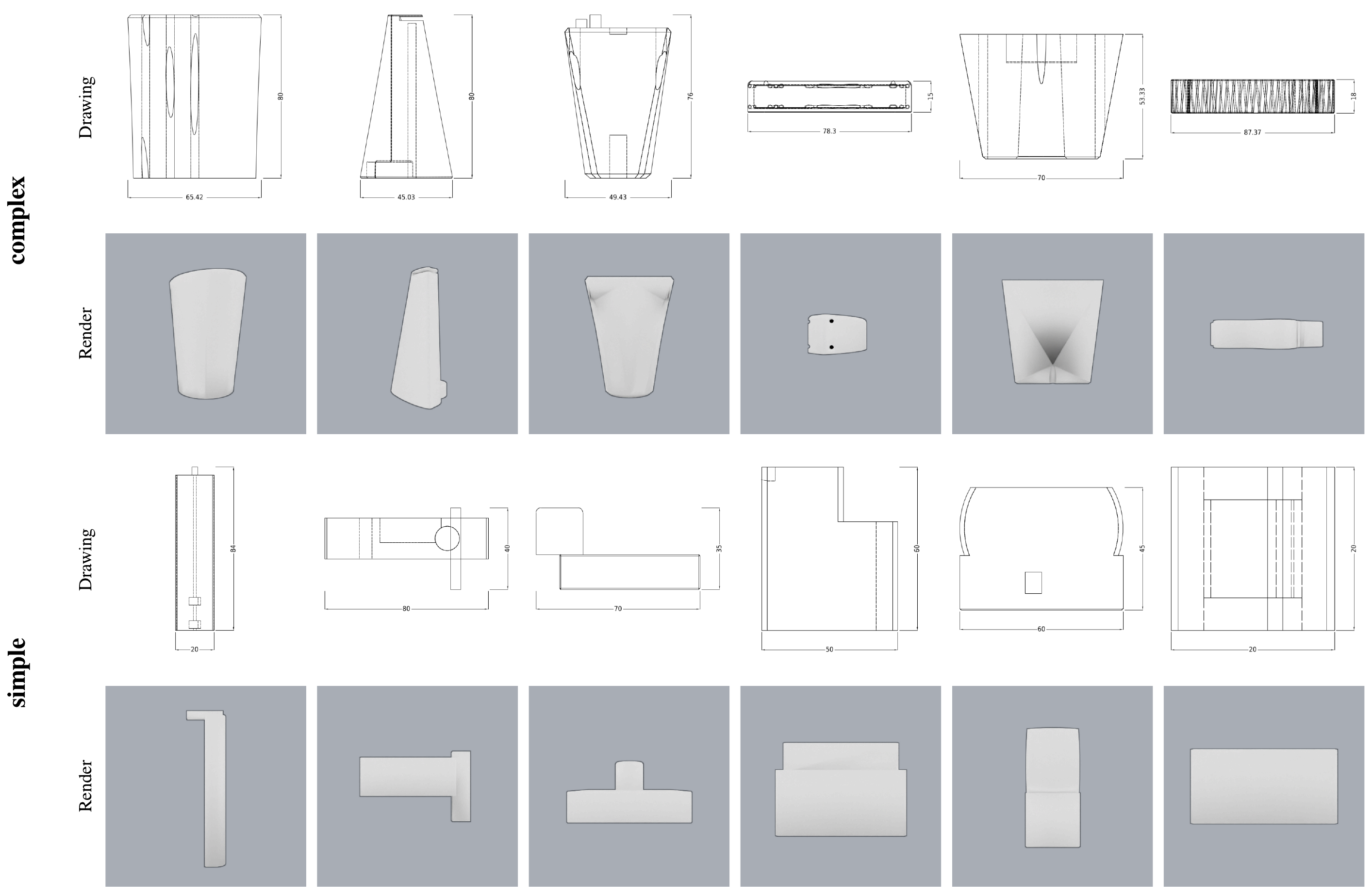}
    \caption{Task 1 examples where all evaluated VLMs misclassified positive pairs.}
    \label{fig:all_failed}
\end{figure*}

\begin{figure*}[p]
    \centering
    \includegraphics[width=\columnwidth]{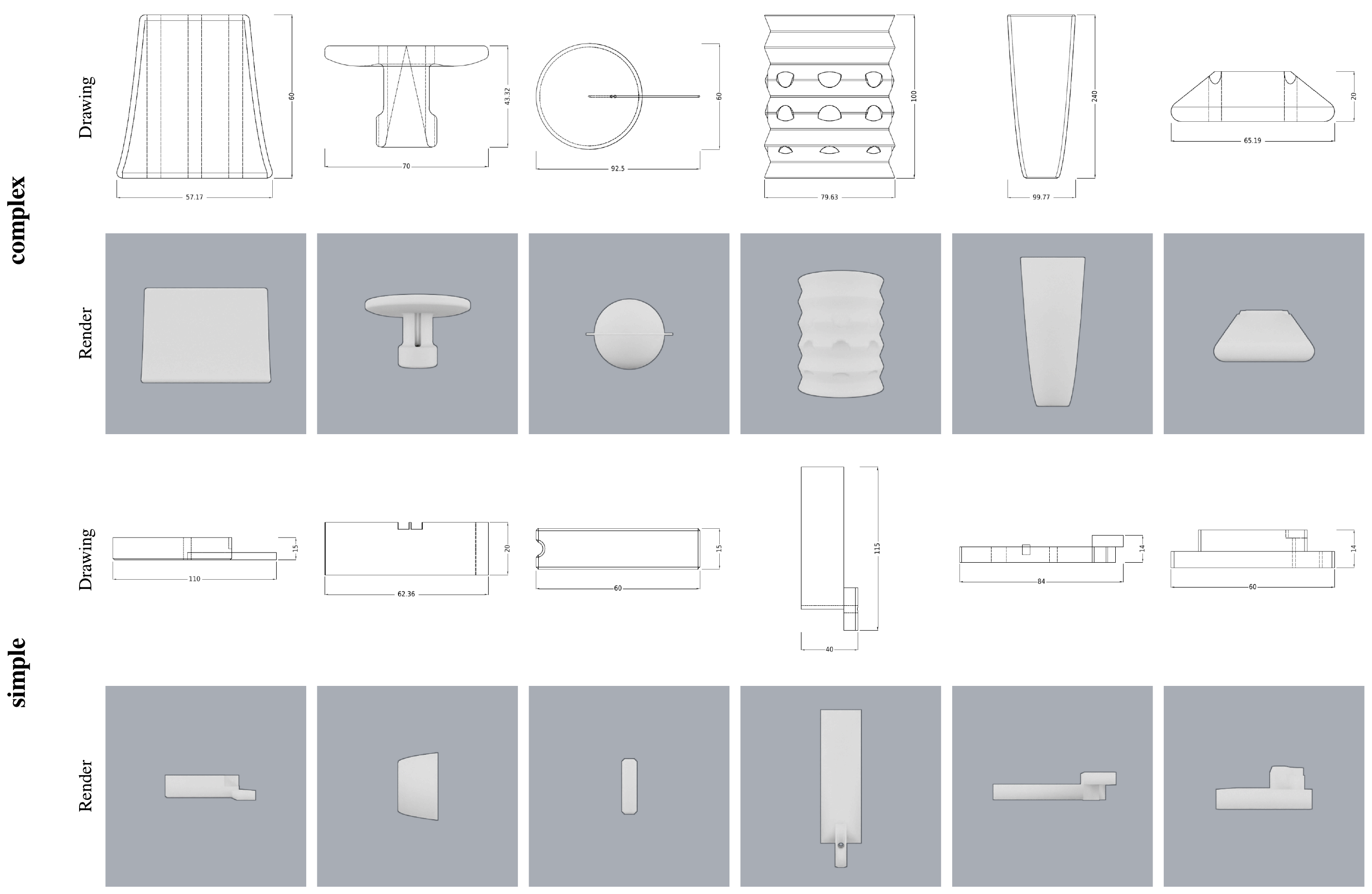}
    \caption{Task 1 examples where all evaluated VLMs correctly matched pairs.}
    \label{fig:types_of_models}
\end{figure*}

\begin{figure*}[p]
    \centering
    \includegraphics[height=0.9\textheight,keepaspectratio]{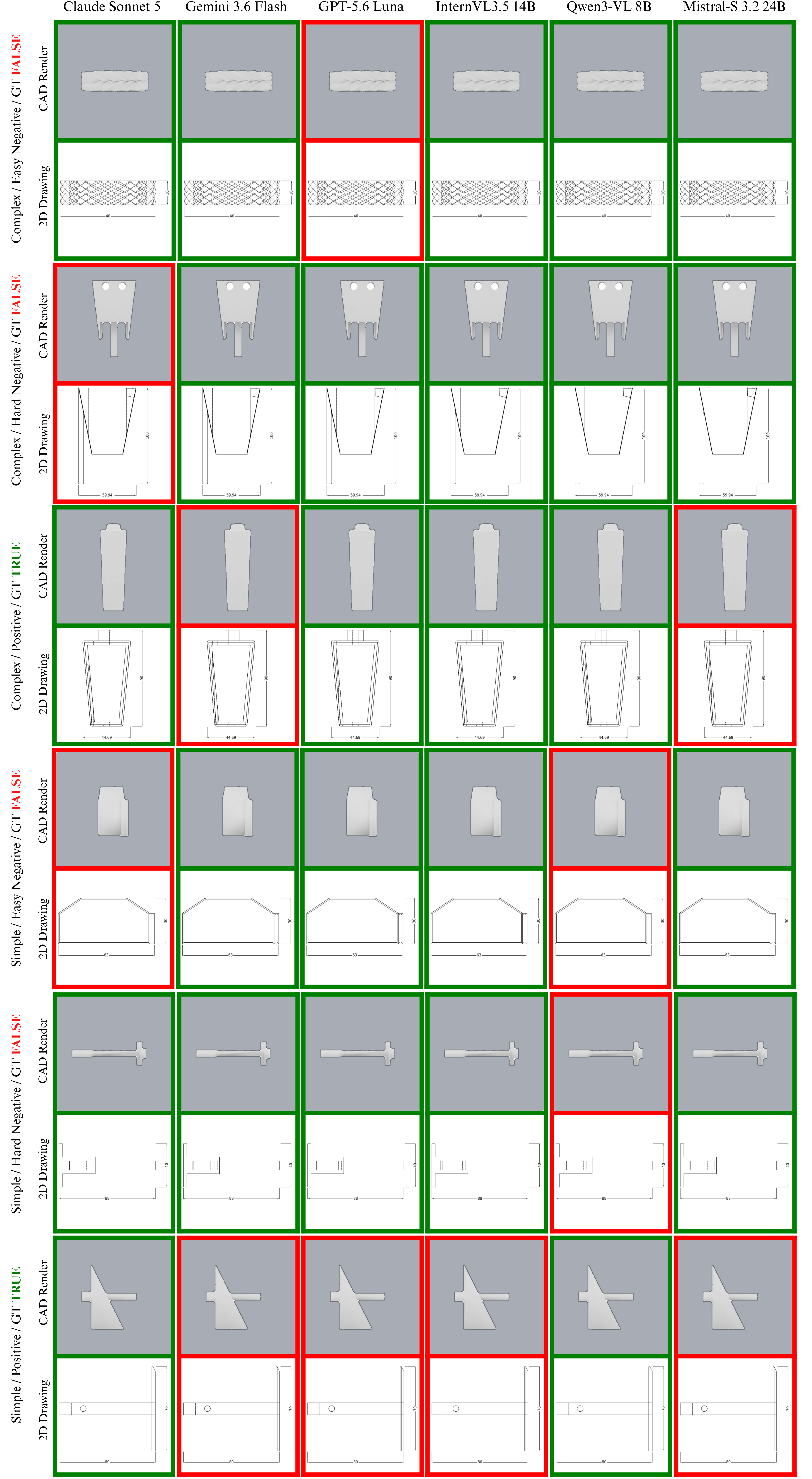}
    \caption{Task 1 qualitative examples showing mixed model predictions. Rows labelled \emph{Hard Negative} are same-class negatives (same prismatic/freeform class) and \emph{Easy Negative} rows are any-class negatives; neither is mined for visual similarity.}
    \label{fig:task1_mixed}
\end{figure*}

Tab.~\ref{tab:task1_failure_analysis} groups Task~1 items by model consensus. Models disagree on $\sim$80\% of items (86--95\% of positives); negatives are almost never failed unanimously ($<1\%$) and are solved by all models in 14--41\% of cases, whereas positives are solved unanimously in only 0.4--3.7\%. Figs.~\ref{fig:all_failed}--\ref{fig:task1_mixed} give examples.

\begin{table}[htbp]
    \centering
    \caption{Task 1 failure analysis by surface-type class, negative type, and prediction outcome.}
    \label{tab:task1_failure_analysis}
    \small
    \begin{tabular}{lllr}
        \toprule
        \rowcolor{gray!20}
        \textbf{Group} & \textbf{Tier} & \textbf{Result} & \textbf{Count (\%)} \\
        \midrule
        \multirow{8}{*}{Complex}
        & \multirow{2}{*}{\makecell[l]{Any-cl. neg. ($n{=}122$)}}
        & All success & 50 (41.0\%) \\
        & & Mixed & 72 (59.0\%) \\ \cmidrule{2-4}
        & \multirow{3}{*}{\makecell[l]{Same-cl. neg. ($n{=}133$)}}
        & All failed & 1 (0.8\%) \\
        & & All success & 53 (39.8\%) \\
        & & Mixed & 79 (59.4\%) \\ \cmidrule{2-4}
        & \multirow{3}{*}{\makecell[l]{Positive ($n{=}245$)}}
        & All failed & 25 (10.2\%) \\
        & & All success & 9 (3.7\%) \\
        & & Mixed & 211 (86.1\%) \\
        \midrule
        \multirow{7}{*}{Simple}
        & \multirow{2}{*}{\makecell[l]{Any-cl. neg. ($n{=}128$)}}
        & All success & 27 (21.1\%) \\
        & & Mixed & 101 (78.9\%) \\ \cmidrule{2-4}
        & \multirow{2}{*}{\makecell[l]{Same-cl. neg. ($n{=}117$)}}
        & All success & 17 (14.5\%) \\
        & & Mixed & 100 (85.5\%) \\ \cmidrule{2-4}
        & \multirow{3}{*}{\makecell[l]{Positive ($n{=}255$)}}
        & All failed & 13 (5.1\%) \\
        & & All success & 1 (0.4\%) \\
        & & Mixed & 241 (94.5\%) \\
        \bottomrule
    \end{tabular}
\end{table}

\section{Fine-Tuning Splits and Pair Construction}
\label{app:finetune}

Data and settings used to adapt Qwen3-VL~8B to Task~1; all steps are deterministic given the released seed (13) and scripts.

\subsection{Splits}
The test split is the \gls{vlm} evaluation set (Sec.~\ref{sec:visual_language_dataset}); the remaining models are split by identifier into training and validation (5\%), and every pair is built within one split (Tab.~\ref{tab:ft_pairs}).

\subsection{Render View Labels}
\label{sec:supp_view_labels}
Each render's view is read from its camera pose by mapping the dominant
axis to the projection directions of Tab.~\ref{tab:cameras}, and verified
on 200 models against the drawings by silhouette IoU. Every render matches
its own axis best (mean IoU 0.85--0.92 against 0.31--0.84 for other views).

\subsection{Hard Negatives}
Hard negatives come from a model-free silhouette descriptor (six $16{\times}16$ view masks plus aspect ratios), so they share the outline and differ in interior structure such as holes, steps, and hidden lines; a nearest neighbour is far closer than a random model (cosine $0.85$ vs.\ $0.01$).

\subsection{Pairs}
Each model contributes two clean drawing queries from two of its six views, each with one positive and one negative; 70\% of negatives are drawn from the three nearest silhouette neighbours, the rest at random, and the render view is sampled independently of whether the pair is positive or negative (Tab.~\ref{tab:ft_pairs}).

\begin{table}[!ht]
\centering
\caption{\textbf{Fine-tuning pairs.} Same view: share of positive / negative pairs whose render shows the drawing's view.}
\label{tab:ft_pairs}
\resizebox{\columnwidth}{!}{%
\begin{tabular}{lrrrrc}
\toprule
\rowcolor{gray!20}
\textbf{Split} & \textbf{Models} & \textbf{Pairs} & \textbf{Pos.} & \textbf{Hard neg.} & \textbf{Same view} \\
\midrule
Train      & 27{,}479 & 109{,}916 & 54{,}958 & 38{,}483 & 0.166 / 0.166 \\
Validation & 1{,}446  & 5{,}784   & 2{,}892  & 2{,}058  & 0.169 / 0.162 \\
\bottomrule
\end{tabular}}
\end{table}

Test pairs are the Task~1 items of the main text ($1{,}000$ test models, front-view drawing; 500 positives, 250 any-class and 250 same-class negatives), in clean and dimensioned form, so zero-shot and fine-tuned rows are scored on identical pairs.

\subsection{Training}
LoRA~\cite{hu2021loralowrankadaptationlarge} adapters cover the language model and the vision encoder, the base model is held in 4-bit during training, and the loss covers only the answer tokens. Tab.~\ref{tab:ft_config} gives the training prompt, which places the drawing first and the render second, and the full configuration.

\begin{table*}[t]
\centering
\caption{\textbf{Fine-tuning prompt and configuration} (Qwen3-VL~8B). The prompt is used for training only; evaluation uses the Task~1 prompt of the main text.}
\label{tab:ft_config}
\begin{minipage}[c]{0.35\textwidth}
\fcolorbox{blue!40!black}{blue!3}{\parbox{\dimexpr\linewidth-2\fboxsep-2\fboxrule\relax}{%
\small\textbf{VLM Prompt.} \itshape You are shown two images of a mechanical part. The first image is a 2D engineering drawing. The second image is a shaded render of a 3D CAD model. The two images may show the part from different viewpoints. Do they depict the same part? Answer with exactly one word: Yes or No.}}%
\end{minipage}\hfill
\begin{minipage}[c]{0.62\textwidth}
\centering\small
\setlength{\tabcolsep}{4pt}
\begin{tabular}{ll}
\toprule
\rowcolor{gray!20}
\textbf{Setting} & \textbf{Value} \\
\midrule
LoRA rank / $\alpha$ / dropout & 16 / 32 / 0.05 \\
Trainable parameters & 52.5M (0.60\%) \\
LoRA targets (LM) & \texttt{q,k,v,o,gate,up,down\_proj} \\
LoRA targets (vision) & \texttt{qkv,proj,linear\_fc1,linear\_fc2} \\
Optimizer & AdamW, lr $10^{-4}$, cosine schedule, 3\% warmup \\
Effective batch & 16 (1 pair $\times$ 16 accumulation) \\
Training pairs & 20{,}000, 1 epoch \\
Validation & 500 pairs every 250 steps, best loss kept \\
Base model & 4-bit NF4 (QLoRA~\cite{dettmers2023qlora}); vision tower, LM head bf16 \\
Images & $512{\times}512$\,px, 256 tokens each \\
Hardware & 1$\times$ L4 GPU (24\,GB) \\
\bottomrule
\end{tabular}
\end{minipage}
\end{table*}
\subsection{Evaluation}
All \emph{Qwen3-VL 8B (FT)} numbers come from the zero-shot harness: the Task~1--3 prompts of Sec.~\ref{sec:vision_language_models}, the same item sets, exact string matching, greedy decoding (rather than the default sampling of the zero-shot open models, Tab.~\ref{tab:vlm_settings}), and the LoRA weights merged into the base model after one epoch (1{,}250 steps at an effective batch of 16, final training loss 0.19). Tasks~2 and~3 were never seen in training and measure transfer. Under the chain-of-thought prompt the model answered with a single word on every item and reproduced its forced-choice predictions exactly, so Table~\ref{tab:cot} repeats those numbers.

\subsection{Caveats}
Hard negatives are selected by outline alone, so they test whether a model reads interior detail rather than whether it separates parts with different outlines but matching local features; and all splits come from one generator, so gains measure in-distribution adaptation (Sec. 7).

\end{document}